%% file: main.tex
\documentclass{article}
\usepackage[final,main]{neurips_2026}
\usepackage[utf8]{inputenc}
\usepackage[T1]{fontenc}
\usepackage{hyperref}
\usepackage{url}
\usepackage{booktabs}
\usepackage{amsfonts}
\usepackage{nicefrac}
\usepackage{microtype}
\usepackage{xcolor}
\usepackage{mathtools}
\usepackage{bm}
\usepackage{graphicx}
\usepackage{placeins}
\usepackage{multirow}
\usepackage{colortbl}
\usepackage{cleveref}  
\usepackage{algorithm}
\usepackage[noend]{algorithmic}
\hypersetup{hypertexnames=false}
\title{Learning Dynamic Evidence Routes for Vision Transformer Probing}

\author{%
  Zice Wang\thanks{Email: wangzc1@mails.neu.edu.cn} \\
  School of Future Technology \\
  Northeastern University \\
  Shenyang, China \\
  \and  
Zhenyu Zhang\thanks{Email: charlie@bit.edu.cn} \\
  School of Computer Science and Technology \\
  Beijing Institute of Technology \\
  Beijing, China \\
}

\begin{document}

\maketitle

\input{sections/abstract}
\input{sections/introduction}

\input{sections/related_work}
\input{sections/method}
\input{sections/experimental_setup}
\input{sections/results}
\input{sections/diagnostic_validation}
\FloatBarrier
\input{sections/discussion}
\input{sections/limitations}
\input{sections/conclusion}

\bibliographystyle{plainnat}
\bibliography{references}

\appendix
\renewcommand{\theHtable}{\thesection.\arabic{table}}
\renewcommand{\theHfigure}{\thesection.\arabic{figure}}
\input{appendix/technical-appendices}

\include{checklist}
\end{document}

%% file: sections/abstract.tex
\begin{abstract}
Probing frozen vision transformers typically uses permutation-invariant aggregation (GAP or $\texttt{[CLS]}$), treating patch tokens as an unstructured set. Content-dependent probes such as self-attention are useful accuracy controls, but they do not expose a fixed token schedule or fixed position weights for auditing.
We introduce \textbf{SSMProbe}, a probe of frozen vision transformers that learns a Sinkhorn patch route and reads it with a diagonal S4 decoder. The decoder's causal linear time-invariant (LTI) recurrence yields fixed position coefficients, exposing how patch arrangement interacts with sequential weighting. Held-out spatial predictability measures route organization beyond a location-bias null. On CUB and Cars, MAE, BEiT, DINOv2, and supervised ViT yield spatially organized routes with distinct readout profiles: DINOv2 is \texttt{[CLS]}-dominant, while MAE benefits from learned routing over fixed scans. On MAE/ImageNet, Sinkhorn reaches 70.33\% accuracy, 5.55 percentage points above the strongest fixed scan.
\end{abstract}

%% file: sections/introduction.tex
\section{Introduction}

Lightweight probes evaluate information accessible in frozen vision transformers (ViTs). Global Average Pooling (GAP) and \texttt{[CLS]} use global representations, while self-attention and deep MLPs learn content-dependent aggregation \citep{dosovitskiy2020image,caron2021emerging}. We study how patch evidence is arranged for sequential readout.

Pre-training objectives such as DINOv2 \citep{oquab2023dinov2}, MAE \citep{he2022masked}, and BEiT \citep{bao2021beit} support different global and patch readouts. An inspectable patch sequence connects these accuracy profiles to the spatial organization of selected evidence.

\textbf{SSMProbe} learns an \emph{evidence route} with a Sinkhorn router and reads it with a diagonal S4 decoder \citep{gu2021efficiently}. S4's native causal linear time-invariant (LTI) recurrence supplies fixed position coefficients after training. The resulting route--readout decomposition supports spatial measurements and analytical sequence perturbations. Appendix~\ref{sec:appendix_complexity} reports parameters and computation budgets.

\begin{figure}[t]
\centering
\includegraphics[width=\textwidth]{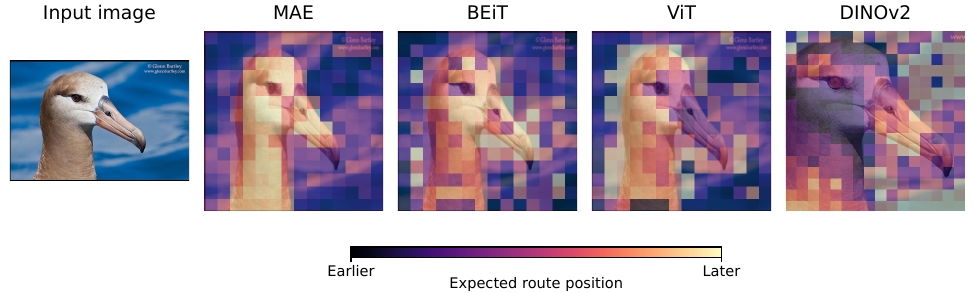}
\caption{\textbf{Patch evidence routes.} The same CUB image is shown alongside the SSMProbe route for four frozen backbones. Color encodes each source patch's normalized expected position in the learned sequence, from earlier (dark) to later (bright). Each overlay uses its backbone's image preprocessing and patch grid. Spatial compactness and bird-foreground alignment are measured by pairwise patch distances and official-mask coverage, respectively.}
\label{fig:route-overview}
\end{figure}

\Cref{fig:route-overview} visualizes routes on a shared image. We quantify their organization by held-out spatial predictability beyond marginal location bias, then measure foreground alignment, seed reproducibility, and sensitivity to sequence perturbations.

On CUB and Cars, all four backbones yield positive held-out spatial information. MAE benefits from learned routing; BEiT and ViT support strong patch aggregation; DINOv2 retains a \texttt{[CLS]}-dominant profile. Decoder controls show that route geometry also depends on the trained readout.

Our contributions are: (1)~\textbf{SSMProbe}, a Sinkhorn+S4 diagnostic that exposes a learned route and fixed positional coefficients; (2)~\textbf{evidence-route geometry}, a compact-vs-dispersed analysis across ViT pre-training objectives; and (3)~\textbf{analytical readout audits} through decomposability, spectrum, perturbation, and skip-path measurements.

%% file: sections/related_work.tex
\section{Related Work}

\textbf{Information Organization in Vision Transformers.} Pre-training objectives structure token information differently \citep{oquab2023dinov2}: contrastive and supervised models often make \texttt{[CLS]} linearly useful \citep{caron2021emerging}, whereas MAE \citep{he2022masked} emphasizes dense patch reconstruction. DeepSets \citep{zaheer2017deepsets} and NetVLAD \citep{arandjelovic2016netvlad} aggregate patch sets; pruning methods \citep{wu2023ppt,bolya2023tome} reduce token-processing costs. SSMProbe measures spatial evidence through an explicit route and fixed positional readout.

\textbf{State Space Models in Vision.} S4 \citep{gu2021efficiently}, Mamba \citep{gu2023mamba}, and visual SSMs \citep{zhu2024vim,liu2024vmamba} are usually studied as efficient backbones. SSMProbe uses S4 to read frozen features: its linear time-invariant form gives an explicit decomposition of ordering effects (\hyperref[prop:decomposability_gap]{Proposition~1}).

\textbf{Probing and Differentiable Routing.} Linear probing evaluates representation quality via feature separability \citep{alain2016understanding}, but ViT patch probes also face an ordering choice. Our router builds on differentiable sorting and Sinkhorn optimal transport \citep{mena2018gumbel,cuturi2019sorting}, including NeuralSort \citep{grover2019stochastic} and SoftSort \citep{prillo2020softsort}. We use these tools to interrogate frozen patch features by ranking them before sequential aggregation.

\textbf{Spatial and Token-Level Structure.} Spatial pretext tasks such as Jigsaw-ViT \citep{chen2023jigsawvit} and PuzzleTuning \citep{zhang2023puzzletuning} explicitly train image representations to recover spatial arrangements. Focal self-attention \citep{yang2021focal} instead enriches ViT attention with local-global interactions. These approaches modify training objectives or backbone computation; SSMProbe complements them as a post-hoc diagnostic of the route induced over frozen patch features by a specified downstream readout.

%% file: sections/method.tex
\section{Method}

\subsection{Problem Setup}
Given an image $x \in \mathbb{R}^{H \times W \times C}$, a frozen Vision Transformer encoder \citep{he2022masked,bao2021beit,oquab2023dinov2,dosovitskiy2020image} (e.g., MAE, BEiT, DINOv2, or ViT) processes it into non-overlapping patches $x_p \in \mathbb{R}^{N \times (P^2 C)}$, where $N = HW/P^2$ for images whose spatial dimensions are divisible by the patch size $P$. The encoder produces final-layer hidden states:
\[
H = [h_{\texttt{cls}}, h_1, \dots, h_N] \in \mathbb{R}^{(N+1) \times d},
\]
where $d$ is the latent dimension. The frozen backbone supplies patch tokens $T=[h_1,\dots,h_N] \in \mathbb{R}^{N \times d}$ to the S4 probe and $h_{\texttt{cls}}$ to the CLS baseline.

\subsection{Structured State Space Sequence Models (S4)}
Our probing mechanism uses a diagonal S4 layer \citep{gu2021efficiently}, a discrete Linear Time-Invariant (LTI) system with state transition $x_k = \overline{\mathbf{A}} x_{k-1} + \overline{\mathbf{B}} u_k$ and output $y_k = \mathbf{C} x_k + \mathbf{D} u_k$. Our \texttt{torch-s4} implementation parameterizes continuous modes as $\lambda_i=-\mathrm{softplus}(a_i)+\mathrm{i}\,\omega_i$ and applies a bilinear transform, placing discrete modes inside the unit disk (Appendix~\ref{sec:appendix_math}). S4 provides a native causal recurrence with a content-independent kernel: ordering effects, spectral attenuation, and the skip path are directly measurable.

\subsection{SSMProbe Head Formulation}
For each sample, we apply a permutation matrix $\mathbf{P} \in \{0, 1\}^{N \times N}$ with $\mathbf{P}\mathbf{1}=\mathbf{1}$ and $\mathbf{P}^\top\!\mathbf{1}=\mathbf{1}$ (or its Sinkhorn soft relaxation) to the token sequence $T$, yielding $\tilde{T} = \mathbf{P}^\top T \in \mathbb{R}^{N \times d}$. We then process the $d$-dimensional features across the sequence length using a diagonal S4 block applied channel-wise:
\[
Z = \mathrm{S4}(\tilde{T}) \in \mathbb{R}^{N \times d}, \quad z_{\text{out}} = Z_N.
\]
The final sequence output $z_{\text{out}}$ aggregates information along the chosen traversal path. A linear projection produces logits $\hat{y} = W z_{\text{out}} + b$. Downstream supervision trains $\theta_{\text{probe}} = \{\mathbf{A}, \mathbf{B}, \mathbf{C}, \mathbf{D}, \Delta, W, b\}$ and the router scorer when present.

\subsection{The SSM as a Diagnostic Instrument}
\label{sec:diagnostic}

\paragraph{Evidence-route position.} Given the Sinkhorn assignment $\mathbf{P}^{(K)}$ defined in Section~\ref{sec:sinkhorn}, the expected S4 position of source patch $i$ is
\[
\tau_i = \sum_{t=1}^{N} t\,\mathbf{P}^{(K)}_{i,t}.
\]
We inspect each patch through this position and the coefficient applied after routing (Proposition~1).

\paragraph{Spatial information in routed evidence.} Let $\mathcal{G}(i)\in\{1,\ldots,\sqrt{N}\}^2$ be the original 2D grid coordinate of patch token $i$, and let $\pi_{1:k}$ be the top-$k$ source-token route induced by the trained scorer. We measure spatial organization by held-out route predictability beyond a location-bias null. We estimate an exchangeable null $p_0(i)\propto m(i)$ from a calibration split, where $m(i)$ is the empirical marginal frequency with which grid position $i$ appears in top-$k$ routes. We compare it to a one-parameter spatial route model
\[
p_{\mathrm{geo}}(i_t\mid i_{<t};\sigma) \propto
m(i_t)\exp\!\left(-\frac{d(i_t,\{i_1,\ldots,i_{t-1}\})^2}{2\sigma^2}\right),
\]
where $d$ is the distance to the nearest previously routed patch; both models sample without replacement. The spatial scale $\sigma$ is selected on the calibration split and evaluated on held-out images. Our primary geometry diagnostic is the excess held-out code length saved by the spatial model,
\[
\mathrm{ERSI}_k = \frac{1}{k\log 2}\left[\log p_{\mathrm{geo}}(\pi_{1:k}) - \log p_0(\pi_{1:k})\right]
\quad\text{bits per routed patch.}
\]
ERSI can be negative on held-out data, but our $\sigma$ grid includes a very large value that makes $p_{\mathrm{geo}}\approx p_0$; reported zeros therefore indicate that the selected spatial model collapses to the null within rounding. Positive values indicate spatial organization beyond marginal location bias, and small fitted $\sigma$ indicates local evidence routes. We report task accuracy alongside ERSI to relate spatial organization to classification.

We also report the pairwise compactness ratio $R_k=D_k/D_k^{\mathrm{rand}}$, where $D_k$ is the mean pairwise grid distance among top-$k$ routed patches. Connected components, convex hull area, minimum-spanning-tree length, and coarse-bin entropy describe complementary aspects of route shape (Appendix~\ref{sec:appendix_route_geometry}).

\paragraph{Controlled interventions.} Fixed scans, random permutations, and reversal measure the decoder's dependence on route scheduling. Cross-seed router swaps measure route reuse. We report within-checkpoint accuracy changes and compare learned routes with deterministic traversals.

\paragraph{Fixed-coefficient decomposition.}
S4's recurrence defines the position coefficients below. A fixed-position linear decoder parameterizes them directly (Appendix~\ref{sec:route-controls}); self-attention uses input-dependent coefficients and offers positional-encoding design choices.

\vspace{2pt}
\noindent\textbf{Proposition 1 (Decomposability gap).}
\phantomsection\label{prop:decomposability_gap}
Let $f_\theta(T, \pi)$ denote the logits produced by sequential decoder $\theta$ on patch tokens $T$ arranged in order $\pi$.
\begin{enumerate}
\item \textbf{S4 decoder.} Since our classifier reads the final hidden/output vector $z_{\text{out}}=Z_N$, the forward pass simplifies to
$f_{\text{S4}}(T, \pi) = \sum_{k=1}^{N} \alpha_k \cdot t_{\pi(k)} + b$,
where the fixed coefficients are
$\alpha_k = W\mathbf{C}\,\bar{A}^{N-k}\bar{B} + \mathbf{1}_{\{k=N\}} W\mathbf{D}$.
Here $W$ is the classifier and $\mathbf{C}$ the S4 state-to-output projection. The direct feedthrough term contributes at the final routed position. The fixed coefficients determine logits for every alternative permutation $\pi'$ analytically.
\item \textbf{Transformer.} A self-attention probe computes 
$f_{\text{TF}}(T, \pi) = g\bigl(\sum_{j} a_j(T, \pi) \cdot (W_V\, t_{\pi(j)})\bigr)$,
where the attention coefficients $a_j(T, \pi)$ depend on the input tokens and their arrangement.
\end{enumerate}
\vspace{2pt}

Analytical permutation logits match direct S4 forwards within $5\times10^{-6}$ absolute error (Appendix~\ref{sec:app-validation-bruteforce}). A position-free Transformer is permutation-equivariant before pooling \citep{zaheer2017deepsets}.

\subsection{Differentiable Sinkhorn Token Routing}
\label{sec:sinkhorn}
We structure token ordering as a 1D optimal transport problem. A lightweight linear scorer assigns each token $h_i$ a standardized scalar score $s_i = \mathbf{w}^\top h_i$, which is matched to evenly spaced rank targets by a squared-distance cost. Sinkhorn-Knopp \citep{cuturi2013sinkhorn} ($K{=}20, \tau{=}0.1$) yields a differentiable doubly-stochastic permutation $\mathbf{P}^{(K)}$ (Appendix~\ref{sec:appendix_sinkhorn}), and $\tilde{T} = {\mathbf{P}^{(K)}}^\top T$. Joint optimization of $\mathbf{w}$ and the probe induces a task-conditioned route. Cross-dataset scorer transfer measures reuse of this learned ranking.

\subsection{Static 2D Traversal Families}
For our fixed-scan baselines, we evaluate deterministic heuristics that linearize the 2D patch grid: simple raster (single row-major), 4-dir VMamba (row/column sweeps following SS2D \citep{liu2024vmamba}), 4-dir diagonal, and 4-dir snake mappings. The 4-dir layout families average their directional variants. \Cref{fig:scan_orders} visualizes these heuristic strategies (detailed definitions in Appendix~\ref{sec:appendix_scans}).

%% file: sections/experimental_setup.tex
\section{Experimental Setup}
\label{sec:experimental-setup}

\noindent\textbf{Backbones and datasets.} We evaluate four frozen ViT backbones spanning distinct pre-training paradigms: MAE (\texttt{facebook/vit-mae-base}), BEiT (\path{microsoft/beit-base-patch16-224-pt22k-ft22k}), DINOv2 (\texttt{facebook/dinov2-base}), and supervised ViT (\texttt{google/vit-base-patch16-224}). Our ImageNet-1K study focuses on MAE, and the cross-backbone comparison uses CUB-200-2011 \citep{wah2011cub} and Stanford Cars \citep{krause20133d} as fine-grained transfer benchmarks.

\medskip
\noindent\textbf{Probe families and optimization.} Beyond GAP and CLS, we compare attention pooling, content-weighted pooling, Top-$k$ pooling, Transformer, DeepSets, Bi-GRU, fixed scans, and the learned Sinkhorn route; full head details are deferred to Appendix~\ref{sec:appendix_implementation}. All heads are trained with AdamW at learning rate $10^{-3}$, cosine schedule, batch size 256, and no weight decay for 5 epochs on ImageNet-1K and 100 epochs on CUB/Cars. The backbone is always frozen, and each head uses an independent optimizer (Appendix~\ref{sec:appendix_math}, Algorithm~\ref{alg:joint_training}).

%% file: sections/results.tex
\section{Results}

\newcommand{\bestacc}[1]{\cellcolor{yellow!30}\textbf{#1}}
\newcommand{\secondacc}[1]{\cellcolor{green!18}\underline{#1}}
\newcommand{\thirdacc}[1]{\cellcolor{blue!12}#1}

\label{sec:diagnostic-results}

We compare readout accuracy, source-grid route geometry, foreground alignment, and decoder dependence across frozen backbones.

\subsection{MAE: Task-Conditioned Evidence Routes}
\label{sec:diagnostic-mae}

\paragraph{ImageNet-1K: route reliability.}
\Cref{tab:mae-main} compares MAE readouts under the five-epoch protocol.

\begin{table}[ht]
\centering
\caption{Frozen MAE probing on ImageNet-1K (five-seed summaries). Scan/permutation and Bi-GRU controls use final checkpoints and source-grid patch order; their error bars are sample standard deviations.}
\label{tab:mae-main}
\begin{tabular}{llc}
\toprule
Tier & Method & Top-1 Acc. (\%) \\
\midrule
Floor & GAP & 58.10 $\pm$ 0.03 \\
& CLS & 56.65 $\pm$ 0.03 \\
\cmidrule{2-3}
Order only & Raster scan & 63.64 $\pm$ 0.05 \\
& 4-dir (VMamba-style) & 64.68 $\pm$ 0.07 \\
& 4-dir (Snake) & 64.76 $\pm$ 0.06 \\
& 4-dir (Diagonal) & 64.78 $\pm$ 0.07 \\
& Random-Fixed Perm + S4 & 63.42 $\pm$ 0.31 \\
& Random-Dynamic Perm + S4 & 63.95 $\pm$ 0.06 \\
\cmidrule{2-3}
Content only & Attention Pool & 67.70 $\pm$ 0.04 \\
& Content-Weighted Pool & 67.68 $\pm$ 0.06 \\
\cmidrule{2-3}
\textbf{Full method} & \textbf{Sinkhorn (learned)} & \bestacc{70.33 $\pm$ 0.05} \\
\cmidrule{2-3}
Ablation & TopK Pool (k=16) & 50.24 $\pm$ 2.57 \\
 & Transformer & 68.10 $\pm$ 0.15 \\
 & DeepSets & \thirdacc{68.48 $\pm$ 0.05} \\
 & Sinkhorn + 1D-Conv & 66.60 $\pm$ 0.07 \\
 & Bi-GRU (Raster) & \secondacc{69.31 $\pm$ 0.12} \\
\cmidrule{2-3}
\textit{Reference} & \textit{Transformer ref. (3M params)} & \textit{71.61 $\pm$ 0.14} \\
\bottomrule
\end{tabular}
\end{table}

\medskip
\noindent
Sinkhorn reaches 70.33\%, gaining 5.55 points over the strongest fixed scan and 6.38 over random-dynamic ordering. Across five seeds, source-grid routes give $\mathrm{ERSI}_{20}=.311\pm.006$ bits/patch and $R_{20}=.581\pm.002$ (Appendix~\ref{sec:imagenet-source-geometry}).

\paragraph{Fine-grained transfer: spatially organized learned routes.}
On CUB and Cars, fixed S4 scans fit the training labels closely but reach 6.7--10.5\% and 5.2--7.0\% test accuracy. Dynamic random ordering reaches 28.4\% and 29.4\%, while Sinkhorn reaches 56.7\% and 54.9\% (\Cref{tab:ablation-fine-grained}). The source-aligned 512-image audits yield positive ERSI and compact routes on both datasets (\Cref{tab:diagnostic-summary}). Thus, MAE supports spatially organized learned routes where deterministic traversals generalize weakly.

\subsection{DINOv2: CLS-Dominant Readout}
\label{sec:diagnostic-dinov2}

On CUB, DINOv2 \citep{oquab2023dinov2} CLS reaches 89.49\% against GAP at 68.55\%, the largest CLS--GAP separation among the four backbones. Fixed scans span 35.45--56.03\%, random-dynamic ordering reaches 68.91\%, and Sinkhorn reaches 82.28\% (\Cref{tab:ablation-fine-grained}).

On Cars, CLS reaches 87.54\% and Sinkhorn 84.83\%, while fixed scans span 23.05--36.83\%. DINOv2 also has the highest CUB ERSI (.484 bits/patch), showing spatially predictable patch routes alongside a CLS-dominant accuracy profile.

\subsection{ViT and BEiT: Compact Patch Evidence}
\label{sec:diagnostic-vit-beit}

\paragraph{ViT (supervised): compact and balanced.} CLS and GAP have similar accuracy on both datasets for supervised ViT \citep{dosovitskiy2020image}. Sinkhorn is close to these readouts on CUB and exceeds them on Cars. Positive ERSI and compactness ratios $R_k=.669$ and $.561$ indicate predictable, localized patch selections (\Cref{tab:diagnostic-summary}).

\paragraph{BEiT: patch-dominant and compact.} BEiT \citep{bao2021beit} has stronger GAP than CLS readouts. Sinkhorn is close to GAP on CUB and exceeds it by 5.80 points on Cars. Its routes have positive ERSI and $R_k\approx.60$ on both datasets, pairing strong patch aggregation with compact evidence selection.

\subsection{Cross-Backbone Diagnostic Comparison}
\label{sec:diagnostic-cross}

\begin{table}[ht]
\centering
\caption{Top-1 accuracy (mean $\pm$ std, \%; five seeds except $\dagger$). MAE, BEiT, and DINOv2 use self-supervised pretraining; ViT is supervised. Colors rank the top three capacity-controlled readouts in each column. The wider Transformer reference appears in its own group.}
\label{tab:ablation-fine-grained}
\begin{tabular}{lcccc}
\toprule
 \multirow{2}{*}{Method} & \multicolumn{2}{c}{MAE} & \multicolumn{2}{c}{BEiT} \\
\cmidrule(lr){2-3} \cmidrule(lr){4-5}
& CUB & Cars & CUB & Cars \\
\midrule
CLS            & 28.88 $\pm$ 0.10 & 16.64 $\pm$ 0.11 & 80.80 $\pm$ 0.13 & 57.28 $\pm$ 0.28 \\
GAP            & 19.40 $\pm$ 0.10 & 17.62 $\pm$ 0.05 & \secondacc{88.27 $\pm$ 0.13} & 76.41 $\pm$ 0.10 \\
\midrule
Raster scan    & 6.72 $\pm$ 0.12 & 5.34 $\pm$ 0.21 & 75.90 $\pm$ 0.16 & 54.64 $\pm$ 0.35 \\
4-dir (Snake)  & 9.84 $\pm$ 0.23 & 6.02 $\pm$ 0.12 & 80.57 $\pm$ 0.40 & 51.65 $\pm$ 0.59 \\
4-dir (Diagonal)& 10.51 $\pm$ 0.20 & 6.97 $\pm$ 0.16 & 80.86 $\pm$ 0.46 & 54.50 $\pm$ 0.40 \\
4-dir (VMamba) & 8.91 $\pm$ 0.09 & 5.21 $\pm$ 0.10 & 80.12 $\pm$ 0.08 & 51.72 $\pm$ 0.21 \\
Random-Fixed Perm + S4 & 6.66 $\pm$ 1.43 & 5.90 $\pm$ 1.72 & 81.68 $\pm$ 0.95 & 58.22 $\pm$ 4.99 \\
Random-Dynamic Perm + S4 & 28.39 $\pm$ 0.28 & 29.41 $\pm$ 0.15 & 87.48 $\pm$ 0.15 & 75.08 $\pm$ 0.22 \\
\midrule
Sinkhorn (learned)& \secondacc{56.67 $\pm$ 7.54} & \secondacc{54.86 $\pm$ 4.35} & \thirdacc{88.12 $\pm$ 0.32} & \bestacc{82.21 $\pm$ 0.64} \\
Transformer & \bestacc{61.03 $\pm$ 0.65} & \bestacc{64.13 $\pm$ 0.57} & 86.03 $\pm$ 0.37 & \secondacc{79.71 $\pm$ 0.42} \\
DeepSets & \thirdacc{34.69 $\pm$ 0.44} & \thirdacc{39.96 $\pm$ 0.44} & \bestacc{88.60 $\pm$ 0.11} & \thirdacc{78.47 $\pm$ 0.22} \\
Sinkhorn + 1D-Conv & 31.57 $\pm$ 0.53 & 33.86 $\pm$ 1.92 & 87.73 $\pm$ 0.20 & 71.64 $\pm$ 0.51 \\
Bi-GRU (Raster) & 18.76 $\pm$ 0.89 & 15.29 $\pm$ 0.72 & 79.15 $\pm$ 0.69 & 27.48 $\pm$ 0.23 \\
\cmidrule{2-5}
\textit{Transformer ref.} & \textit{70.08 $\pm$ 0.57} & \textit{71.11 $\pm$ 0.65} & \textit{88.03 $\pm$ 0.14} & \textit{85.71 $\pm$ 0.17} \\
\midrule
 \multirow{2}{*}{Method} & \multicolumn{2}{c}{ViT} & \multicolumn{2}{c}{DINOv2} \\
\cmidrule(lr){2-3} \cmidrule(lr){4-5}
& CUB & Cars & CUB & Cars \\
\midrule
CLS            & \bestacc{85.45 $\pm$ 0.10} & \thirdacc{57.01 $\pm$ 0.21} & \bestacc{89.49 $\pm$ 0.21} & \bestacc{87.54 $\pm$ 0.11} \\
GAP            & \thirdacc{85.14 $\pm$ 0.13} & 56.89 $\pm$ 0.10 & 68.55 $\pm$ 0.23 & 75.13 $\pm$ 0.09 \\
\midrule
Raster scan    & 79.90 $\pm$ 0.24 & 40.51 $\pm$ 0.21 & 35.45 $\pm$ 0.32 & 23.05 $\pm$ 0.39 \\
4-dir (Snake)  & 82.17 $\pm$ 0.21 & 41.85 $\pm$ 0.18 & 56.03 $\pm$ 0.22 & 36.64 $\pm$ 0.40 \\
4-dir (Diagonal)& 82.32 $\pm$ 0.16 & 43.90 $\pm$ 0.15 & 50.71 $\pm$ 0.21 & 32.97 $\pm$ 0.46 \\
4-dir (VMamba) & 81.73 $\pm$ 0.17 & 41.84 $\pm$ 0.16 & 50.40 $\pm$ 0.15$^{\dagger}$ & 36.83 $\pm$ 0.59 \\
Random-Fixed Perm + S4 & 81.36 $\pm$ 0.58 & 44.48 $\pm$ 2.08 & 37.24 $\pm$ 5.49 & 35.52 $\pm$ 6.50 \\
Random-Dynamic Perm + S4 & 83.85 $\pm$ 0.20 & 55.24 $\pm$ 0.29 & 68.91 $\pm$ 0.18 & 74.20 $\pm$ 0.14 \\
\midrule
Sinkhorn (learned)& \secondacc{85.22 $\pm$ 0.19} & \bestacc{62.38 $\pm$ 0.15} & \thirdacc{82.28 $\pm$ 1.60} & \thirdacc{84.83 $\pm$ 0.60} \\
Transformer & 81.22 $\pm$ 0.26 & \secondacc{58.40 $\pm$ 0.63} & \secondacc{83.56 $\pm$ 0.19} & \secondacc{85.97 $\pm$ 0.46} \\
DeepSets & 84.03 $\pm$ 0.25 & 56.98 $\pm$ 0.34 & 76.62 $\pm$ 0.29 & 81.06 $\pm$ 0.16 \\
Sinkhorn + 1D-Conv & 83.88 $\pm$ 0.30 & 55.34 $\pm$ 0.35 & 71.19 $\pm$ 0.45 & 74.24 $\pm$ 0.52 \\
Bi-GRU (Raster) & 82.12 $\pm$ 0.46 & 40.87 $\pm$ 0.52 & 59.76 $\pm$ 0.80 & 50.52 $\pm$ 1.07 \\
\cmidrule{2-5}
\textit{Transformer ref.} & \textit{85.17 $\pm$ 0.07} & \textit{68.70 $\pm$ 0.39} & \textit{87.31 $\pm$ 0.15} & \textit{89.84 $\pm$ 0.16} \\
\bottomrule
\end{tabular}
\par\smallskip
\begin{minipage}{\linewidth}
\footnotesize
S4 four-direction controls use seeds 42--46 and final checkpoints, with sample standard deviations. Each path covers all 196 MAE/BEiT/ViT patches or 256 DINOv2 patches. MAE scan/permutation and Bi-GRU controls restore source-grid order; Bi-GRU uses Raster order. $\dagger$: four valid seeds and one numerical failure (seed 44, NaN loss at epoch 96), retained without replacement and excluded from the accuracy summary. Appendix~\ref{sec:appendix_scans} specifies the paths.
\end{minipage}
\end{table}

\Cref{tab:diagnostic-summary} pairs route geometry with readout accuracy; Appendix~\ref{sec:appendix_route_geometry} defines the grid descriptors.

\begin{table}[ht]
\centering
\caption{Trained evidence-route geometry. $\mathrm{ERSI}_{20}$ measures held-out spatial-information gain over the location-bias null (bits/patch); $\sigma$ is the fitted spatial scale and $R_k$ the pairwise compactness ratio. Accuracy provides readout reliability context.}
\label{tab:diagnostic-summary}
\small
\setlength{\tabcolsep}{4.5pt}
\begin{tabular}{lcccc}
\toprule
Measure & MAE & BEiT & ViT & DINOv2 \\
\midrule
Spatial information $\mathrm{ERSI}_{20}$ (CUB) & $0.318$ & $0.175$ & $0.113$ & $0.484$ \\
Spatial information $\mathrm{ERSI}_{20}$ (Cars) & $0.336$ & $0.276$ & $0.296$ & $0.328$ \\
Fitted $\sigma$ (CUB / Cars) & $3$ / $2$ & $3$ / $2$ & $3$ / $2$ & $3$ / $3$ \\
Compactness ref. $R_k$ (CUB / Cars) & $0.687$ / $0.567$ & $0.609$ / $0.599$ & $0.669$ / $0.561$ & $0.697$ / $0.680$ \\
Sinkhorn accuracy (CUB) & $56.67$ & $88.12$ & $85.22$ & $82.28$ \\
Sinkhorn accuracy (Cars) & $54.86$ & $82.21$ & $62.38$ & $84.83$ \\
\midrule
\textbf{Characteristic pattern} & Organized & Organized & Organized & Organized,\\
& routing-sensitive & compact patch & compact local & CLS-leaning \\
\bottomrule
\end{tabular}
\end{table}

All four backbones yield positive ERSI and $R_k<1$ on both datasets. Their differences lie in readout profiles and the degree of spatial organization.

\paragraph{Object foreground.}
Official CUB masks connect route geometry to bird foreground (\Cref{tab:cub-foreground}). All four backbones exceed both random selection and the location-bias null. MAE reaches 55.40\% foreground coverage against 12.60\% random coverage and a 33.25\% location-bias null, an excess of 22.15 percentage points.

\begin{table}[ht]
\centering\small
\caption{Top-20 routed patches on bird foreground (\%), over all 5,794 CUB test images. Mask preprocessing matches each backbone. DINOv2 uses 256 patches; the others use 196.}
\label{tab:cub-foreground}
\begin{tabular}{lrrrr}
\toprule
Selection & MAE & BEiT & ViT & DINOv2 \\
\midrule
Sinkhorn route & 55.40 & 43.09 & 32.73 & 58.81 \\
Uniform random & 12.60 & 12.60 & 12.60 & 21.64 \\
Location-bias null & 33.25 & 31.43 & 26.60 & 31.83 \\
\bottomrule
\end{tabular}
\end{table}

\begin{figure}[!htbp]
\centering
\includegraphics[width=.88\textwidth]{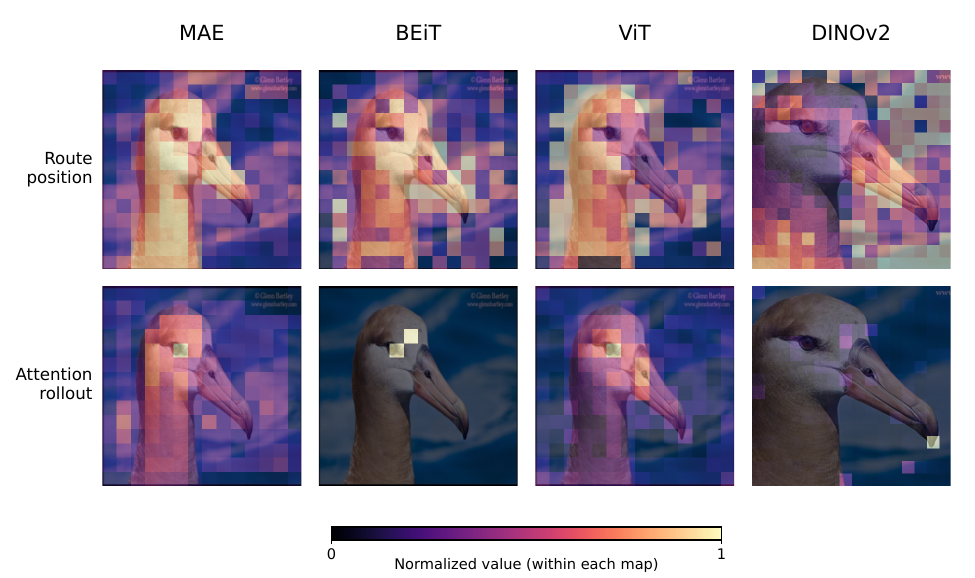}
\caption{Same-image CUB comparison. Top: normalized expected route position (later positions are hotter). Bottom: attention rollout. Each column uses the corresponding backbone's preprocessed image and full patch grid; MAE patches retain source-grid order. Colors are normalized within each map.}
\label{fig:route-attention-comparison}
\end{figure}

\paragraph{Route and backbone attention.}
\Cref{fig:route-attention-comparison} compares expected route positions with backbone attention rollout on the same CUB image. The route is learned for the downstream S4 readout. Rollout multiplies head-averaged layer attention with the residual correction $\tfrac12 A_\ell+\tfrac12 I$, then normalizes the CLS row over patches. \Cref{tab:route-attention-agreement} reports agreement on the first 256 CUB test images. MAE rollout has mean patch variance $1.094\times10^{-6}$, indicating a relatively flat relevance field; its rank agreement is interpreted alongside this variation.

\begin{table}[ht]
\centering\small
\caption{Mean per-image agreement between Sinkhorn scores and backbone attention on 256 shared CUB test images. Heads are averaged. MAE/BEiT/ViT use 196 patches; DINOv2 base uses 256 patches and zero register tokens. Top-20 IoU is intersection size divided by union size.}
\label{tab:route-attention-agreement}
\begin{tabular}{lcccc}
\toprule
& \multicolumn{2}{c}{Rollout} & \multicolumn{2}{c}{Last-layer CLS} \\
\cmidrule(lr){2-3}\cmidrule(lr){4-5}
Backbone & Spearman & Top-20 IoU & Spearman & Top-20 IoU \\
\midrule
MAE & .013 & .160 & .185 & .233 \\
BEiT & .201 & .193 & .526 & .296 \\
ViT & .129 & .129 & .380 & .066 \\
DINOv2 & .081 & .109 & .111 & .167 \\
\bottomrule
\end{tabular}
\end{table}

%% file: sections/diagnostic_validation.tex
\section{Evidence-Route Validation}
\label{sec:diagnostic-validation}

\paragraph{Decoder dependence.}
A learned image-independent scalar $\alpha_k$ at each routed position gives the fixed-weight readout $W\sum_k\alpha_k\tilde h_k+b$. On DINOv2/CUB it reaches $80.62\pm1.11\%$, compared with $81.86\pm1.23\%$ for S4 under a matched five-seed protocol. MAE fixed weights reach $46.49\pm18.17\%$, with substantial seed variation. \Cref{tab:decoder-geometry-summary} quantifies route geometry: BEiT is compact in five of five S4 seeds and three of five fixed-weight seeds; DINOv2 is compact in all five seeds with either decoder. S4's causal LTI recurrence and the directly parameterized fixed coefficients induce different strengths of spatial concentration. Appendix~\ref{sec:route-controls} gives per-seed results and training-family conventions.

\begin{table}[ht]
\centering\small
\caption{Decoder-dependent top-20 compactness on the same 256 CUB test images. Mean $\pm$ sample std over seeds 42--46; the final column counts seeds with $R_{20}<1$ (smaller pairwise grid distance than random patches). These controls use training families distinct from \Cref{tab:diagnostic-summary}.}
\label{tab:decoder-geometry-summary}
\begin{tabular}{llcc}
\toprule
Backbone & Decoder & $R_{20}$ & Compact seeds \\
\midrule
MAE & Fixed weights & $.772\pm.173$ & 5/5 \\
BEiT & Fixed weights & $.806\pm.246$ & 3/5 \\
DINOv2 & Fixed weights & $.834\pm.115$ & 5/5 \\
BEiT & S4 & $.639\pm.071$ & 5/5 \\
DINOv2 & S4 & $.530\pm.074$ & 5/5 \\
\bottomrule
\end{tabular}
\end{table}

\paragraph{Ordering and readout interventions.}

Across five MAE/ImageNet checkpoints, own-router and cross-seed-router accuracies are 71.18\% and 71.21\%. A separate ordering study gives 71.21\% with the learned route and 34.70\% after scrambling. On a 512-image subset, pairwise route Spearman is $.994$ and top-20 overlap is 94.3\%; reversal reduces accuracy by 66.3 points, while adjacent swaps reduce it by 0.2. These interventions distinguish global sequencing from local changes. The diagonal S4 has stable modes ($\max_i|\lambda_i|=.98996\pm.00030$) and a broad coefficient profile ($n_{90}=105.0\pm2.1$ positions). On the 512-image subset, dropping 20 Sinkhorn-ranked tokens reduces accuracy by 19.9 points, compared with 3.1 for CLS-cosine selection and 0.9 for random selection. Appendix~\ref{sec:appendix_validation} and Appendix~\ref{sec:appendix_results} report each protocol, full spectral statistics, saliency comparisons, and perturbation tables.

%% file: sections/discussion.tex
\section{Discussion}

Learned ordering improves MAE task accuracy over deterministic traversals, linking sequence utility to both patch selection and positional weighting.

Fixed weights retain most DINOv2 accuracy, while BEiT compactness varies more across seeds than with S4. A shared decoder exposes cross-backbone differences in route geometry; decoder swaps measure geometry's dependence on positional weighting.

CUB masks quantify object-foreground selection. Coordinate-aligned perturbations show stable route ranks under mild noise and larger changes under cropping; attention comparisons measure agreement with backbone relevance maps (Appendix~\ref{sec:route-controls}).

%% file: sections/limitations.tex
\section{Limitations}
\label{sec:limitations}

\paragraph{Task and decoder dependence.} The linear Sinkhorn scorer is learned jointly with the downstream readout. Its route reflects frozen features, task supervision, and decoder coefficients. BEiT's five-seed compactness changes from $0.639\pm0.071$ under S4 to $0.806\pm0.246$ under fixed weights, including two fixed-weight seeds above the random-distance baseline. Cross-dataset scorer transfer retains reusable signal but falls 12--21\,pp below full retraining (Appendix~\ref{sec:appendix_transfer}).

\paragraph{Evaluation scope.} Experiments cover classification on ImageNet-1K, CUB, and Cars, with object-mask alignment on CUB. Extension to dense prediction and other downstream tasks requires task-specific evaluation. Router hyperparameter sweeps cover accuracy sensitivity; geometry sensitivity across routing configurations remains an open evaluation question.

\paragraph{Input sensitivity.} On MAE/CUB, Gaussian noise of standard deviation .005 yields route Spearman $.995$ and top-20 overlap $96.91\%$, while accuracy changes from $59.38\%$ to $57.81\%$ on 128 images. A crop retaining 90\% of each dimension yields Spearman $.783$ and top-20 overlap $69.30\%$ measured by bilinear coordinate alignment over 144 common source-grid centers (Appendix~\ref{sec:input-control}).

%% file: sections/conclusion.tex
\section{Conclusion}
SSMProbe combines Sinkhorn routing and a causal S4 decoder to measure task-conditioned patch arrangement and fixed positional weighting. Across CUB and Cars, spatially organized routes yield distinct readout profiles. Object masks, spatial nulls, decoder controls, and input perturbations ground this route--readout analysis.

%% file: appendix/technical-appendices.tex
\input{appendix/theory-and-method-details}
\input{appendix/extended-results}
\input{appendix/validation-and-reproducibility}
\input{appendix/route-controls}

%% file: appendix/theory-and-method-details.tex
\section{Theory and Method Details}
\label{sec:appendix_theory}

\subsection{S4 Dynamics and Routed Position Coefficients}
\label{sec:appendix_math}
\label{sec:appendix_s4}

The diagonal S4 maps each input channel through a continuous-time state $x(t)\in\mathbb{R}^{Q}$:
\begin{align}
x'(t) &= \mathbf{A}x(t)+\mathbf{B}u(t), \\
y(t) &= \mathbf{C}x(t)+\mathbf{D}u(t).
\end{align}
The continuous modes are initialized from a stable S4/HiPPO-inspired spectrum~\citep{gu2020hippo,gu2021efficiently} and parameterized as
\[
\lambda_i = -\mathrm{softplus}(a_i) + \mathrm{i}\,\omega_i,
\]
with negative real parts. The bilinear discretization is
\[
\bar{\mathbf{A}}=(I-\Delta\mathbf{A}/2)^{-1}(I+\Delta\mathbf{A}/2),\qquad
\bar{\mathbf{B}}=(I-\Delta\mathbf{A}/2)^{-1}\Delta\mathbf{B},
\]
placing the discrete modes inside the unit disk. Our \texttt{torch-s4} implementation uses $d=768$, $Q=128$, and dropout 0. The learnable step size is initialized at .05 and clamped to $[10^{-3},10^{-1}]$; $\mathbf{D}$ is initialized to zero. Channels share S4 parameters.

For a routed sequence $\tilde{T}=\mathbf{P}^{\top}T$ of length $L$, the discrete recurrence and final emitted output are
\begin{align}
x_k &= \bar{\mathbf{A}}x_{k-1}+\bar{\mathbf{B}}\tilde{t}_k, \\
z_L &= \sum_{k=1}^{L}\mathbf{C}\bar{\mathbf{A}}^{L-k}\bar{\mathbf{B}}\tilde{t}_k+\mathbf{D}\tilde{t}_L.
\end{align}
With classifier $Wz_L+b$, define the fixed position coefficient
\[
\alpha_k=W\mathbf{C}\bar{\mathbf{A}}^{L-k}\bar{\mathbf{B}}+
\mathbf{1}_{\{k=L\}}W\mathbf{D}.
\]
For the full route ($L=N$), substituting $\tilde{t}_k=\sum_i P_{i,k}t_i$ gives
\[
f(T,\mathbf{P})=\sum_{i=1}^{N}\left(\sum_{k=1}^{N}P_{i,k}\alpha_k\right)t_i+b.
\]
The router assigns patch features to these coefficients through downstream supervision. Slow modes retain evidence over longer horizons; faster modes attenuate earlier contributions. Spectral, skip-path, and sequence perturbation audits measure this interaction.

\paragraph{Ordering properties.} GAP satisfies $f(\Pi T)=f(T)$ for any permutation matrix $\Pi$; CLS uses the global token. The S4 decoder generally changes when its routed inputs are permuted because $\alpha_k$ varies with position. The full Sinkhorn+S4 probe canonicalizes input permutations through its router. Accordingly, sequence perturbations are applied after routing to measure decoder order dependence.

\subsection{Joint Training Protocol with Gradient Isolation}
Probe heads share one frozen-backbone forward per batch and use independent optimizers:

\begin{algorithm}
\caption{Joint Training with Per-Head Gradient Isolation}
\label{alg:joint_training}
\begin{algorithmic}
\STATE \textbf{Input:} frozen backbone $B$, training set $\mathcal{D}$, list of probe heads $\mathcal{H}$
\FOR{each head $h \in \mathcal{H}$}
  \STATE optimizer[$h$] $\leftarrow$ AdamW($h$.parameters(), lr=$10^{-3}$)
\ENDFOR
\FOR{epoch $= 1$ \TO $N_{\text{epochs}}$}
  \FOR{(images, labels) $\in \mathcal{D}$}
    \STATE features $\leftarrow$ \texttt{detach}($B$(images)) \COMMENT{no-grad, frozen}
    \FOR{head $h \in \mathcal{H}$}
      \STATE logits[$h$] $\leftarrow h$(features)
      \STATE loss[$h$] $\leftarrow$ CrossEntropy(logits[$h$], labels)
      \STATE loss[$h$].backward() \COMMENT{gradient isolated per head}
      \STATE optimizer[$h$].step()
      \STATE optimizer[$h$].zero\_grad()
    \ENDFOR
  \ENDFOR
\ENDFOR
\end{algorithmic}
\end{algorithm}

Each head uses AdamW (lr=$10^{-3}$, cosine schedule) on its own parameters; detached backbone features are shared across heads.

\subsection{Full Sinkhorn OT Formulation}
\label{sec:appendix_sinkhorn}

The per-token scorer $s_i = \mathbf{w}^\top h_i$ is a bias-free \texttt{Linear}$(d,1)$ layer with default PyTorch initialization. Scores are standardized within each image:
\[
\tilde{s}_i = \frac{s_i - \mu(s)}{\sigma(s) + \epsilon}.
\]
Canonical positions are $p_j=j/(N-1)$ for $j\in\{0,\ldots,N-1\}$, with cost $\mathbf{C}_{i,j}=(\tilde{s}_i-p_j)^2$. Joint supervision of the scorer and S4 readout yields task-conditioned patch ordering; scorer transfer measures reuse across tasks.

The entropy-regularized OT plan:
\[
\mathbf{P}_\tau = \mathop{\mathrm{argmin}}_{\mathbf{P} \in \mathbb{R}_{+}^{N \times N}} \langle \mathbf{P}, \mathbf{C} \rangle - \tau \mathcal{H}(\mathbf{P}) \quad \text{s.t.} \quad \mathbf{P} \mathbf{1} = \mathbf{1}, \; \mathbf{P}^\top \mathbf{1} = \mathbf{1},
\]
where $\mathcal{H}(\mathbf{P})=-\sum_{i,j}\mathbf{P}_{i,j}\log\mathbf{P}_{i,j}$ is elementwise entropy for a matrix of total mass $N$. Initialize $\mathbf{G}=\exp(-\mathbf{C}/\tau)$ and iterate:
\begin{align}
  \mathbf{u}^{(r+1)} &= \mathbf{1} \oslash (\mathbf{G} \mathbf{v}^{(r)}), \\
  \mathbf{v}^{(r+1)} &= \mathbf{1} \oslash (\mathbf{G}^\top \mathbf{u}^{(r+1)}),
\end{align}
After $R{=}20$ Sinkhorn iterations, $\mathbf{P}^{(R)} = \mathrm{diag}(\mathbf{u}^{(R)}) \mathbf{G} \mathrm{diag}(\mathbf{v}^{(R)})$ acts as a soft permutation with temperature $\tau{=}0.1$.

\subsection{Formal Definitions for Evidence-Route Spatial Geometry}
\label{sec:appendix_route_geometry}

This section defines the spatial-information diagnostic and the random-null-calibrated descriptive statistics used in \Cref{tab:diagnostic-summary}. Let the $N$ patch tokens form a square grid of side length $G=\sqrt{N}$. Let $\gamma(i)=(r_i,c_i)\in\{0,\ldots,G-1\}^2$ denote the original 2D coordinate of source token $i$. For a trained route score or contribution field, let
\[
\mathcal{T}_k \subset \{1,\ldots,N\}, \qquad |\mathcal{T}_k|=k,
\]
be the set of top-$k$ source-token indices. All metrics are computed per image and averaged over the evaluation subset. For each statistic $m(\mathcal{T}_k)$, we also sample random top-$k$ sets $\mathcal{R}_k$ uniformly without replacement from the same $G\times G$ grid and report a ratio to the random-null expectation when appropriate:
\[
\rho_m = \frac{m(\mathcal{T}_k)}{\mathbb{E}_{\mathcal{R}_k}[m(\mathcal{R}_k)]}.
\]
For metrics where larger values mean more concentration (largest connected component), compactness corresponds to $\rho_m>1$; for metrics where smaller values mean more concentration (distance, component count, hull area, MST length, entropy), compactness corresponds to $\rho_m<1$.

\paragraph{Evidence-route spatial information.}
Let $\pi_{1:k}=(i_1,\ldots,i_k)$ be the ordered top-$k$ route for one image. We split evaluation images into calibration and held-out subsets. Calibration routes provide a smoothed marginal location prior $m(i)$ capturing dataset-level and probe-level location bias. The exchangeable null samples without replacement according to
\[
p_0(i_t\mid i_{<t}) = \frac{m(i_t)\mathbf{1}\{i_t\notin i_{<t}\}}{\sum_{j\notin i_{<t}}m(j)}.
\]
The spatial alternative uses the same marginal prior but additionally favors locations near previously routed evidence:
\[
p_{\mathrm{geo}}(i_t\mid i_{<t};\sigma) =
\frac{m(i_t)\exp\{-d(i_t,i_{<t})^2/(2\sigma^2)\}\mathbf{1}\{i_t\notin i_{<t}\}}
{\sum_{j\notin i_{<t}}m(j)\exp\{-d(j,i_{<t})^2/(2\sigma^2)\}},
\]
where $d(i_t,i_{<t})=\min_{q<t}\|\gamma(i_t)-\gamma(i_q)\|_2$. For $t=1$, the spatial factor is omitted, so both models use the same marginal prior. We select $\sigma$ from a grid on the calibration split and report the held-out log-likelihood gain
\[
\mathrm{ERSI}_k = \frac{1}{Mk\log 2}\sum_{m=1}^{M}\left[
\log p_{\mathrm{geo}}(\pi^{(m)}_{1:k};\hat{\sigma}) - \log p_0(\pi^{(m)}_{1:k})
\right],
\]
in bits per routed patch over $M$ held-out images. Positive values indicate spatial predictability beyond marginal location bias; negative values indicate lower held-out likelihood than the null. The grid includes a large-$\sigma$ endpoint approximating $p_0$, so selection of this endpoint yields a gain near zero.

\paragraph{Mean pairwise grid distance.}
The descriptive compactness ratio uses
\[
D_k(\mathcal{T}_k)=\frac{2}{k(k-1)}\sum_{\substack{i,j\in\mathcal{T}_k\\ i<j}}\|\gamma(i)-\gamma(j)\|_2,
\qquad
R_k=\frac{D_k(\mathcal{T}_k)}{\mathbb{E}_{\mathcal{R}_k}[D_k(\mathcal{R}_k)]}.
\]
Values $R_k\approx1$ indicate random-like dispersion; values below one indicate that routed evidence is closer than random top-$k$ patches.

\paragraph{4-neighbor connected components.}
Construct an undirected graph $\mathcal{G}_k=(\mathcal{T}_k,E_k)$ with edges between horizontally or vertically adjacent selected patches:
\[
E_k=\bigl\{(i,j): i,j\in\mathcal{T}_k,\; \|\gamma(i)-\gamma(j)\|_1=1\bigr\}.
\]
Let $\mathcal{C}_1,\ldots,\mathcal{C}_{M_k}$ be the connected components of $\mathcal{G}_k$. We report the largest-component fraction
\[
L_k=\frac{1}{k}\max_{1\le q\le M_k}|\mathcal{C}_q|
\]
and the number of components $M_k$. A compact route should have larger $L_k$ and smaller $M_k$ than a random top-$k$ set.

\paragraph{Convex hull area.}
Let $\mathrm{Hull}(\mathcal{T}_k)$ be the Euclidean convex hull of the point set $\{(c_i,r_i):i\in\mathcal{T}_k\}\subset\mathbb{R}^2$. We normalize its polygon area by the full grid area:
\[
A_k=\frac{\mathrm{area}\bigl(\mathrm{Hull}(\mathcal{T}_k)\bigr)}{(G-1)^2}.
\]
Smaller $A_k$ indicates that top-ranked evidence occupies a smaller spatial envelope.

\paragraph{Euclidean minimum-spanning-tree length.}
Let $\mathcal{K}_k$ be the complete graph over selected coordinates with edge weight $w_{ij}=\|\gamma(i)-\gamma(j)\|_2$. Let $\mathrm{MST}(\mathcal{T}_k)$ denote its Euclidean minimum spanning tree. We use the normalized tree length
\[
S_k=\frac{1}{\sqrt{k}(G-1)}\sum_{(i,j)\in\mathrm{MST}(\mathcal{T}_k)}\|\gamma(i)-\gamma(j)\|_2.
\]
This graph-length statistic is less sensitive to a single outlier pair than the full pairwise distance while still measuring spatial spread.

\paragraph{Entropy over coarse spatial bins.}
Partition the $G\times G$ grid into $B\times B$ coarse bins $\{\mathcal{B}_b\}_{b=1}^{B^2}$; in our audit $B=4$. Define
\[
p_b=\frac{1}{k}\sum_{i\in\mathcal{T}_k}\mathbf{1}\{\gamma(i)\in\mathcal{B}_b\}.
\]
The normalized spatial-bin entropy is
\[
H_k=-\frac{1}{\log(B^2)}\sum_{b=1}^{B^2} p_b\log(p_b+\epsilon),
\]
with a small $\epsilon$ only for numerical stability. Lower entropy indicates that routed evidence occupies fewer coarse image regions.

\paragraph{Descriptive robustness checks.}
The descriptive audit uses the random-null thresholds $R_k<0.9$, $L_k/\mathbb{E}[L_k^{\mathrm{rand}}]>1.5$, $M_k/\mathbb{E}[M_k^{\mathrm{rand}}]<0.85$, $A_k/\mathbb{E}[A_k^{\mathrm{rand}}]<0.85$, $S_k/\mathbb{E}[S_k^{\mathrm{rand}}]<0.9$, and $H_k/\mathbb{E}[H_k^{\mathrm{rand}}]<0.9$ to compare concentration across geometric descriptors.

\subsection{Fixed Traversal Families}
\label{sec:appendix_scans}

We evaluate four families of deterministic scan orders designed to linearize the 2D grid of patches (assumed to form an $\sqrt{N} \times \sqrt{N}$ grid):

\paragraph{Raster scan.} A single row-major traversal (left-to-right, top-to-bottom).

\paragraph{4-dir (VMamba).} Four scans across rows and columns, including row-major forward and reverse, and column-major forward and reverse~\citep{liu2024vmamba}. This follows the SS2D (2D Selective Scan) strategy introduced in VMamba.

\paragraph{4-dir (Diagonal).} Concatenate anti-diagonals in increasing order of $r+c$, visiting rows in increasing order within each anti-diagonal. A horizontal reflection gives a second complete traversal. Each traversal and its reverse form the four paths.

\paragraph{4-dir (Snake).} Alternating directions at each boundary (e.g., left-to-right for even rows, right-to-left for odd rows), performed both row-wise and column-wise.

\begin{figure}[ht]
    \centering
    \includegraphics[width=0.85\textwidth]{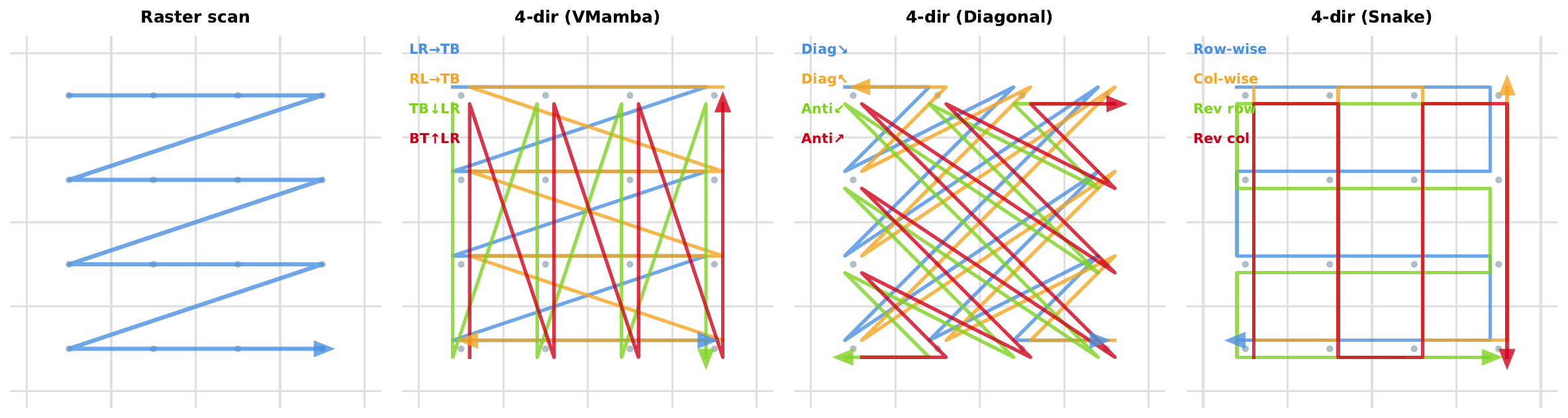}
    \caption{Four scan strategies: (1) raster (row-major), (2) 4-dir VMamba (row/column sweeps), (3) 4-dir diagonal, (4) 4-dir snake (alternating directions).}
    \label{fig:scan_orders}
\end{figure}

For the MAE results in \Cref{tab:mae-main,tab:ablation-fine-grained}, source-grid patch order is restored with \texttt{ids\_restore}. Each four-direction path visits all $N=196$ patches once. The same S4 layer processes each path, and its four final states are averaged before classification. Raster uses one complete row-major path. Four complete paths process $4N$ tokens per image.

BEiT and supervised ViT use four complete paths of 196 patches; DINOv2 uses four complete paths of 256 patches. Each path is a permutation of the full source grid. The four paths share one S4 layer and average their final states before classification, giving 784 patch visits for MAE/BEiT/ViT and 1,024 for DINOv2.

\subsection{Probe Implementation Details}
\label{sec:appendix_implementation}

\paragraph{Checkpoint and reporting protocol.} Each head trains for the full fixed budget: 5 epochs on ImageNet-1K and 100 epochs on CUB/Cars. We report accuracy from the final checkpoint at the end of training. Intermediate evaluations track training progress.

\paragraph{Seeding.} All random seeds (Python, NumPy, PyTorch CUDA) set to a common value per run. Data order fixed by the dataset loader's default shuffling with a seeded generator.

\paragraph{Baselines.} (i)~\textbf{Attention pooling}: single learned query token attending to all patches via dot-product attention (single head, no FFN). (ii)~\textbf{Content-weighted pooling}: softmax token-wise weights from the per-token linear scorer without reordering. (iii)~\textbf{Top-$k$ pooling}: average of top-$k$ highest-scored tokens. These disentangle dynamic token weighting from dynamic token ordering.

\paragraph{Bi-GRU.} The fine-grained Bi-GRU baseline receives source-grid-ordered patches in Raster order. It uses a single bidirectional GRU with 128 hidden units per direction and a linear classifier on the concatenated terminal states.

%% file: appendix/extended-results.tex
\section{Extended Results}
\label{sec:appendix_results}

\paragraph{Protocol map.}
\Cref{tab:appendix-protocol-map} identifies the comparison and checkpoint family for each study. Accuracy changes are compared within each family.

\begin{table}[ht]
\centering
\caption{Result families and evaluation settings.}
\label{tab:appendix-protocol-map}
\small
\begin{tabular}{ll}
\toprule
Table & Setting \\
\midrule
\Cref{tab:mae-main} & MAE/ImageNet main comparison \\
\Cref{tab:ablation-fine-grained} & CUB/Cars cross-backbone main comparison \\
\Cref{tab:routing_ablation} & Separate MAE/CUB routing family \\
\Cref{tab:sinkhorn-grid} & Separate MAE/CUB temperature/iteration sweep \\
\Cref{tab:state-scaling} & Separate MAE/CUB state-size sweep \\
\Cref{tab:scramble-ordering} & Separate ImageNet ordering-sensitivity family \\
\bottomrule
\end{tabular}
\end{table}

\subsection{Layer-wise Representation Evolution}
\begin{figure}[ht]
    \centering
    \includegraphics[width=\textwidth]{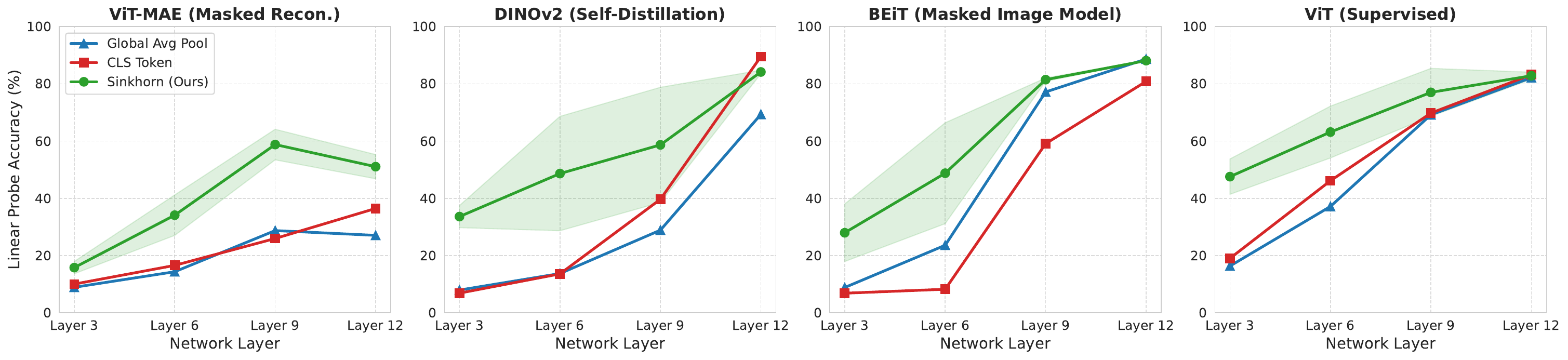}
    \caption{Layer-wise evaluation of probing architectures. We extract token representations at intermediate layers (3, 6, 9, 12) of four distinct Vision Transformer families and evaluate them using Sinkhorn, CLS-based, and GAP-based linear probes. Sinkhorn routing (green) often extracts useful patch evidence earlier than CLS-based probing (red), indicating that route-based readouts can reveal intermediate-layer patch structure.}
    \label{fig:layer-wise-evolution}
\end{figure}

We evaluate Sinkhorn, CLS, and GAP readouts at layers 3, 6, 9, and 12 of frozen supervised ViT, DINOv2, BEiT, and MAE on CUB-200-2011 (\Cref{fig:layer-wise-evolution}).

At DINOv2 layer 3, Sinkhorn reaches 33.62\% against 6.87\% for CLS, showing task-useful patch evidence at intermediate depth. MAE reaches 58.79\% with Sinkhorn against 36.46\% with CLS in this layerwise protocol. Sinkhorn accuracy generally increases with depth across the four backbones.

\subsection{Routing Validation and Ordering Dependence}

We compare Sinkhorn, NeuralSort, and SoftSort on frozen MAE/CUB features, holding the linear scorer, S4 decoder, and optimization settings fixed.

\begin{figure}[ht]
    \centering
    \includegraphics[width=0.9\linewidth]{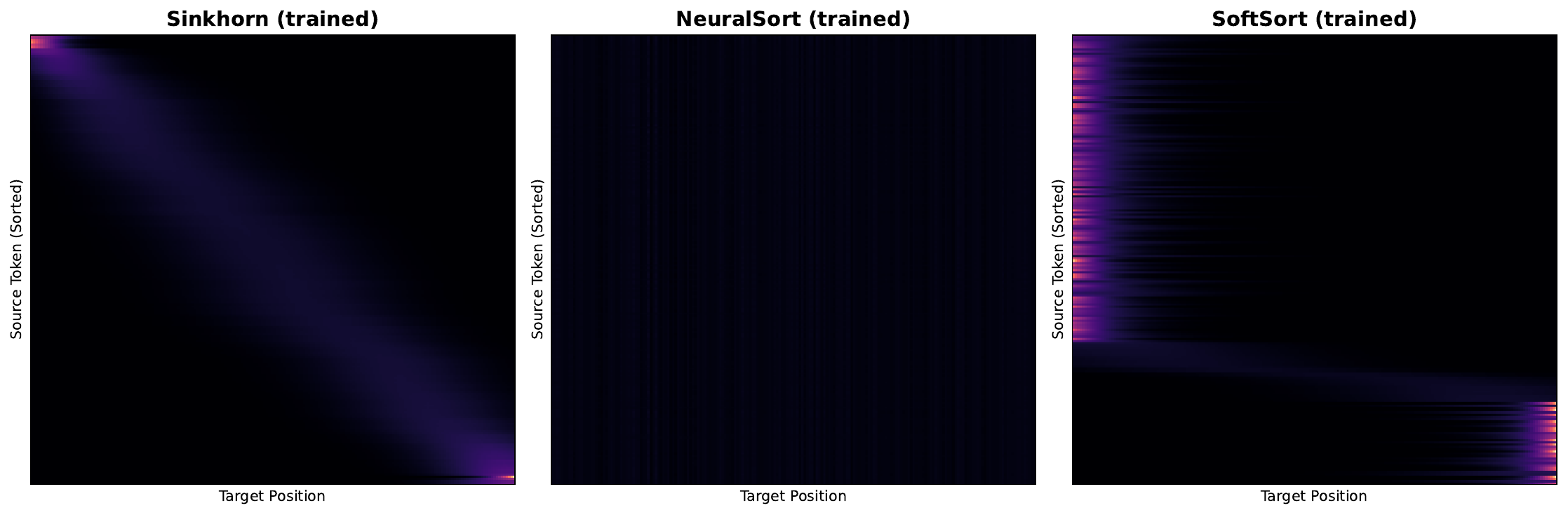}
    \caption{Differentiable assignment matrices, with tokens sorted by each method's argmax target position. Sinkhorn distributes assignments across ranks; NeuralSort has low target coverage; SoftSort concentrates mass near boundary ranks.}
    \label{fig:permutation_matrices}
\end{figure}

\begin{table}[ht]
\centering
\caption{Differentiable routing ablation on frozen MAE/CUB features.}
\label{tab:routing_ablation}
\begin{tabular}{lclc}
\toprule
Routing Mechanism & Method Type & Perm. Invariant? & Top-1 Acc. (\%) \\
\midrule
GAP & Global Pooling & \checkmark & 19.57 \\
CLS & Token Extraction & \checkmark & 29.01 \\
\midrule
NeuralSort & Deterministic Soft-Rank & \texttimes & 29.51 \\
SoftSort & Continuous Sorting & \texttimes & 54.07 \\
Sinkhorn (Ours) & Doubly-Stochastic OT & \texttimes & \textbf{60.32} \\
\bottomrule
\end{tabular}
\end{table}

\medskip
\noindent
\textbf{Readout accuracy.} Sinkhorn reaches 60.32\%, SoftSort 54.07\%, and NeuralSort 29.51\%, compared with GAP at 19.57\% and CLS at 29.01\% (\Cref{tab:routing_ablation}).

\medskip
\noindent
\textbf{Assignment coverage.} We measure the fraction of target positions selected by row-wise argmax, the normalized entropy of their histogram, and boundary mass (\Cref{fig:permutation_matrices}). Sinkhorn covers 112/196 positions (.571) with entropy .868 and row-max p95 .047. NeuralSort selects one position (.005 coverage, entropy 0), with mean row maximum .0077. SoftSort covers 24/196 positions (.122), with entropy .235, mass .883 in the outer 10\% of ranks, and row-max p95 .127.

For the ImageNet ordering study, we randomly permute the soft-routed sequence $\tilde{X}=\mathbf{P}^{\top}X$ before S4. \Cref{tab:scramble-ordering} reports five-seed results for this checkpoint family.

\begin{table}[ht]
\centering
\caption{ImageNet ordering sensitivity: five-seed accuracy and percentage-point change from the learned route.}
\label{tab:scramble-ordering}
\begin{tabular}{lcc}
\toprule
Configuration & ImageNet-1K Acc. & Drop from Normal \\
\midrule
Normal routing (Sinkhorn) & $71.21\pm0.02\%$ & -- \\
Scramble after routing & $34.70\pm0.60\%$ & $-36.51\%$ \\
No routing (S4 on raw tokens) & $21.12\pm0.78\%$ & $-50.09\%$ \\
Random permutation before S4 & $20.80\pm0.41\%$ & $-50.41\%$ \\
\bottomrule
\end{tabular}
\end{table}

Scrambling reduces accuracy by 36.51 percentage points. Raw-token and random-permutation inputs reach 21.12\% and 20.80\%, respectively, measuring the trained decoder's dependence on the learned sequence.

\subsection{Probe Parameters and Computational Budgets}
\label{sec:appendix_complexity}

\Cref{tab:probe-complexity-full} reports exact parameter counts and forward-operation estimates obtained from the implemented probe heads using \texttt{fvcore} JIT tracing with custom handlers for FFTs, complex arithmetic, Sinkhorn normalization, and GRU gates. Profiling uses batch size one, 196 patch tokens of width 768, 1,000 output classes, and evaluation mode; the frozen backbone is excluded. We count a real multiply--add as two operations and a complex multiply--accumulate as eight, with a radix-2 real-FFT estimate of $2.5L\log_2L$ per transform and inverse normalization counted separately. Pointwise arithmetic and reductions are included; indexing, sorting, and memory traffic are excluded. S4 constructs its complex convolution kernel on each forward pass. Four complete paths share parameters, make 784 patch visits, and classify their averaged final states once. \Cref{tab:mae-source-head-cost} separately reports measured latency. The main Transformer row uses width 96; Transformer ref. uses width 768.

\begin{table}[ht]
\centering
\caption{Exact probe parameter counts and estimated forward FLOPs for 196 patch tokens of width 768 and 1,000 output classes (batch size one; frozen backbone excluded). Four-path scans cover the complete grid on each path.}
\label{tab:probe-complexity-full}
\begin{tabular}{lrr}
\toprule
Probe Head & Params (M) & Est. FLOPs (M) \\
\midrule
GAP (linear) & 0.769 & 1.69 \\
CLS (linear) & 0.769 & 1.54 \\
Attention Pool & 0.770 & 2.14 \\
Content-Weighted Pool & 0.770 & 2.14 \\
\cmidrule{2-3}
S4 Raster (state dim 128) & 0.967 & 184.97 \\
VMamba / Diagonal / Snake (four paths) & 0.967 & 735.28 \\
Sinkhorn (S4 + scorer, $I{=}20$) & 0.968 & 250.63 \\
Sinkhorn (S4 + scorer, $I{=}10$) & 0.968 & 247.55 \\
Sinkhorn (S4 + scorer, $I{=}5$) & 0.968 & 246.01 \\
\midrule
Transformer & 0.992 & 75.70 \\
Transformer ref. & 3.430 & 1170.05 \\
DeepSets & 1.066 & 118.53 \\
Random-Fixed Perm + S4 & 0.967 & 184.97 \\
Random-Dynamic Perm + S4 & 0.967 & 184.97 \\
Sinkhorn + 1D-Conv & 0.931 & 355.43 \\
Bi-GRU & 0.947 & 271.41 \\
\bottomrule
\end{tabular}
\end{table}

\subsection{Sinkhorn Hyperparameter Grid Search}
\label{sec:appendix_sinkhorn_grid}

We sweep Sinkhorn iterations $I\in\{1,5,10,20\}$ and temperature $\tau\in\{.1,.2,.5,1.0\}$ on frozen MAE/CUB features (\Cref{tab:sinkhorn-grid}). Five-seed mean accuracy spans 59.79--61.71\% at $\tau\in\{.1,.2\}$ and 49.86--55.70\% at $\tau=1$. The default $I=20$, $\tau=.1$ gives the highest mean in this grid.

\begin{table}[ht]
\centering
\caption{Sinkhorn iteration and temperature sweep on frozen MAE/CUB features (five-seed mean $\pm$ std).}
\label{tab:sinkhorn-grid}
\begin{tabular}{lcccc}
\toprule
 & \multicolumn{4}{c}{Temperature $\tau$} \\
\cmidrule{2-5}
Sinkhorn Iterations $I$ & 0.1 & 0.2 & 0.5 & 1.0 \\
\midrule
1  & 60.12 $\pm$ 0.86 & 60.98 $\pm$ 0.72 & 59.76 $\pm$ 1.57 & 55.70 $\pm$ 1.11 \\
5  & 59.92 $\pm$ 1.85 & 60.64 $\pm$ 2.15 & 56.79 $\pm$ 2.07 & 49.86 $\pm$ 1.56 \\
10 & 60.82 $\pm$ 1.03 & 60.22 $\pm$ 1.29 & 54.91 $\pm$ 2.62 & 50.04 $\pm$ 1.37 \\
20 & \textbf{61.71 $\pm$ 1.00} & 59.79 $\pm$ 0.71 & 56.81 $\pm$ 1.94 & 50.24 $\pm$ 1.64 \\
\bottomrule
\end{tabular}
\end{table}

\subsection{Token Sensitivity Analysis}
\label{sec:appendix_token_sensitivity}

We evaluate seed-42 final MAE/ImageNet probe checkpoints on the first 2,048 validation images. The backbone uses bfloat16 with masking disabled, restores source-grid order with \texttt{ids\_restore}, and supplies float32 features to the heads. Complete four-path traversals retain the relative order of their remaining source patches after removal. Random removal uses per-image seed-42 subsets nested across $k$.

\begin{figure}[ht]
    \centering
    \includegraphics[width=.96\linewidth]{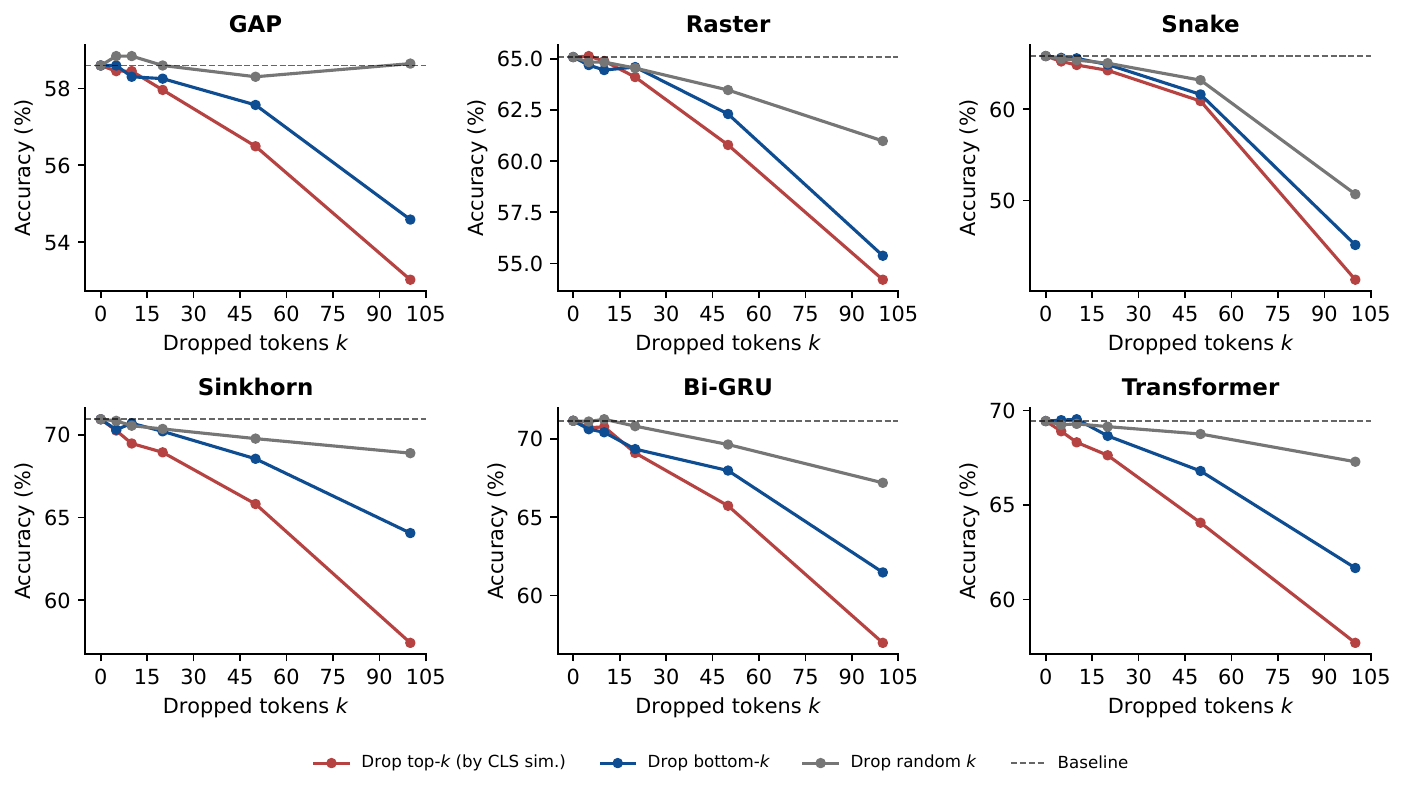}
    \caption{Token dropping on 2,048 MAE/ImageNet validation images, using seed-42 final checkpoints. Six panels show GAP, Raster, complete-path Snake, Sinkhorn, Raster Bi-GRU, and the width-96 Transformer. Top/bottom ranking uses CLS cosine similarity; random removal uses per-image subsets. Dashed lines show intact-token accuracy on the same images.}
    \label{fig:drop_curves}
\end{figure}

\paragraph{Token dropping.} We rank patches by CLS cosine and remove the top-$k$, bottom-$k$, or random $k$ tokens (\Cref{fig:drop_curves}). At $k=100$, top-ranked removal reduces Raster, Snake, Sinkhorn, Bi-GRU, and Transformer accuracy by 10.89, 24.51, 13.53, 14.16, and 11.72 percentage points, respectively. Random removal reduces the same readouts by 4.10, 15.14, 2.05, 3.96, and 2.15 points. Bottom-ranked removal gives intermediate decreases of 9.72, 20.70, 6.88, 9.67, and 7.76 points.

\begin{figure}[ht]
    \centering
    \includegraphics[width=.96\linewidth]{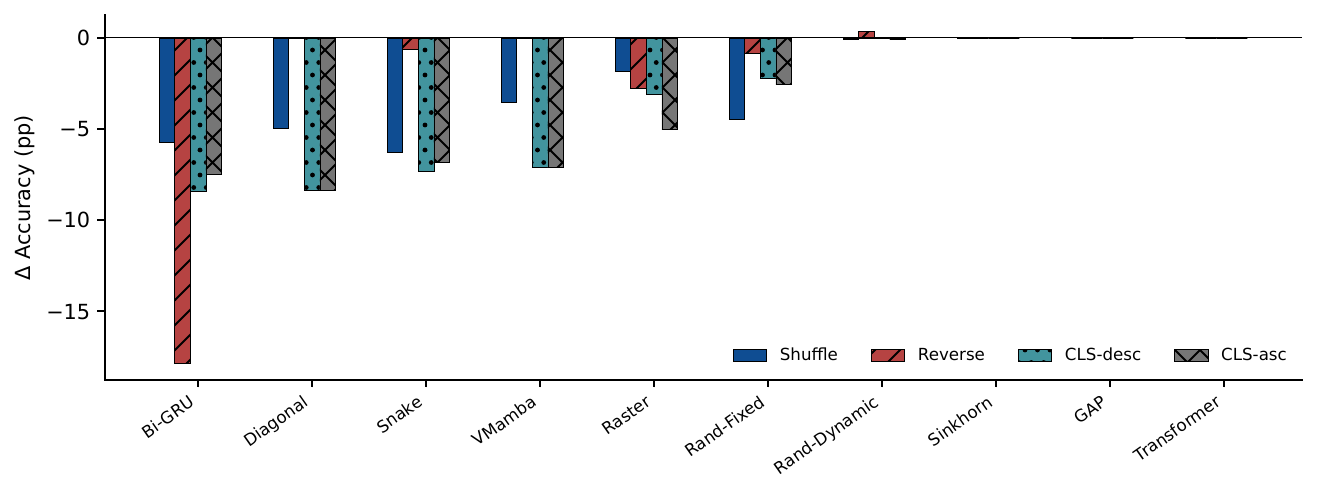}
    \caption{Input-order sensitivity for ten MAE/ImageNet readouts on the same 2,048 validation images. Bars show accuracy changes in percentage points relative to each seed-42 final checkpoint's intact-token baseline. Four-direction scans use complete source-grid paths; Transformer denotes the width-96 control.}
    \label{fig:perm_delta}
\end{figure}

\paragraph{Token permutation.} We shuffle, reverse, or reorder tokens by ascending/descending CLS similarity before inference (\Cref{fig:perm_delta}). Shuffling reduces Raster, Snake, and Bi-GRU accuracy by 1.86, 6.30, and 5.76 points; reversal changes them by $-2.78$, $-.63$, and $-17.87$ points. GAP, Sinkhorn, and the Transformer retain their accuracy under all four input reorderings. VMamba-style and diagonal path sets are closed under reversal, so their averaged readout preserves accuracy under that intervention. Random-dynamic order is sampled anew with a reset seed for each intervention.

\subsection{Cross-Dataset Scorer Transfer}
\label{sec:appendix_transfer}

We freeze the bias-free MAE scorer $s_i=\mathbf{w}^{\top}h_i$ learned on CUB or Cars and train the S4 decoder and classifier on the other dataset. \Cref{tab:cross-transfer} compares both transfer directions with full scorer/decoder training and a frozen random scorer.

\begin{table}[ht]
\centering
\caption{Cross-dataset scorer transfer on frozen MAE features. ``Transfer'': scorer loaded from source checkpoint, only S4+classifier retrained. ``Full retrain'': all parameters (including scorer) trained from scratch on target. ``Random'': random scorer direction, frozen.}
\label{tab:cross-transfer}
\begin{tabular}{llcc}
\toprule
Source & Target & Scorer & Test Acc.\ (\%) \\
\midrule
CUB   & Cars & Transfer (CUB frozen) & 41.2 \\
CUB   & Cars & Full retrain          & 53.4 \\
CUB   & Cars & Random (frozen)       & 29.7 \\
\midrule
Cars  & CUB  & Transfer (Cars frozen) & 40.2 \\
Cars  & CUB  & Full retrain           & 61.2 \\
Cars  & CUB  & Random (frozen)        & 20.8 \\
\bottomrule
\end{tabular}
\end{table}

Transfer exceeds the random scorer by 11.5 points on Cars and 19.4 on CUB, while falling 12.2 and 21.0 points below full training. The 768-parameter scorer retains task-useful signal across both directions, with a measurable benefit from target-task adaptation.

\begin{figure}[ht]
    \centering
    \begin{minipage}{0.90\linewidth}
        \centering
        \includegraphics[width=\linewidth]{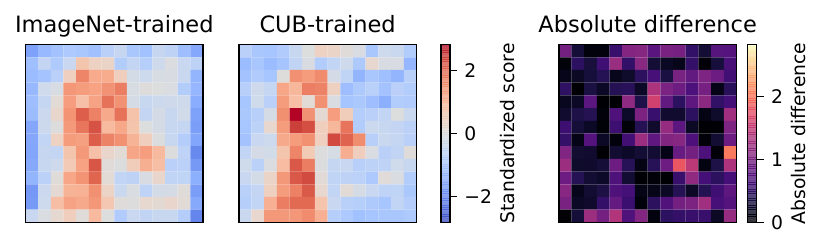}
        \vspace{-4pt}
        \centerline{\small CUB sample 000}
    \end{minipage}

    \medskip
    \begin{minipage}{0.90\linewidth}
        \centering
        \includegraphics[width=\linewidth]{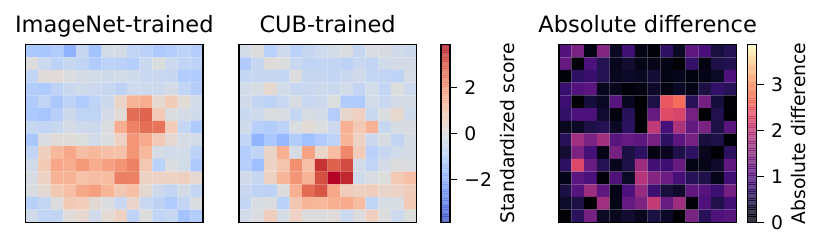}
        \vspace{-4pt}
        \centerline{\small CUB sample 001}
    \end{minipage}

    \caption{Task-adapted scorer heatmaps on two fixed CUB test images, with MAE patches restored to source-grid order. Each triplet shows standardized patch scores from an ImageNet-trained scorer, a CUB-trained scorer, and their absolute difference. Signed fields share a diverging scale; absolute differences use a separate nonnegative scale.}
    \label{fig:scorer-task-adaptation}
\end{figure}

\begin{figure}[ht]
    \centering
    \includegraphics[width=.90\linewidth]{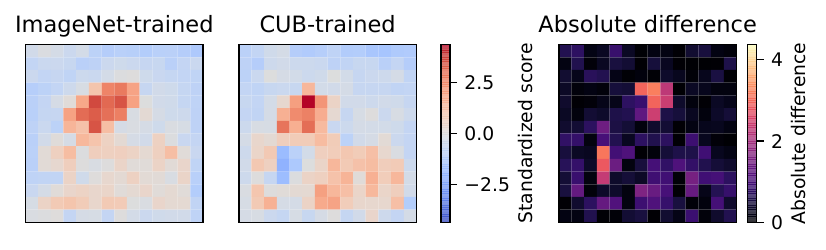}
    \centerline{\small ImageNet validation sample 000}
    \medskip
    \includegraphics[width=.90\linewidth]{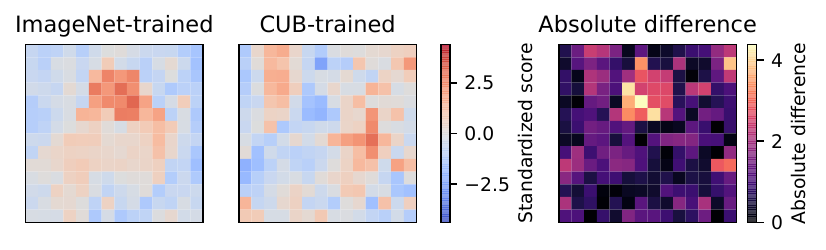}
    \centerline{\small ImageNet validation sample 001}
    \caption{Task-adapted scorer fields on the first two ImageNet validation images, using source-aligned MAE patches. Each row compares the ImageNet-trained and CUB-trained scorers on the same image, followed by the absolute difference. Signed fields share a diverging scale within each row; differences use a separate nonnegative scale.}
    \label{fig:scorer-task-adaptation-imagenet}
\end{figure}

\subsection{State-Size Scaling}
\label{sec:appendix_state_scaling}

We vary S4 state dimension on frozen MAE/CUB features with Sinkhorn settings $I=20$, $\tau=.1$. Mean accuracy rises from 55.22\% at dimension 1 to 60.42\% at dimension 4, then spans 59.19--60.79\% across dimensions 8--256 (\Cref{tab:state-scaling}).

\begin{table}[ht]
\centering
\caption{S4 state-size sweep on frozen MAE/CUB features with Sinkhorn routing ($I=20$, $\tau=.1$; five-seed mean $\pm$ std).}
\label{tab:state-scaling}
\begin{tabular}{lcc}
\toprule
State Dim.\ $N$ & Top-1 Acc.\ (\%) & $\Delta$ vs.\ GAP \\
\midrule
1   & 55.22 $\pm$ 1.72 & +35.65 \\
2   & 57.51 $\pm$ 2.07 & +37.94 \\
4   & 60.42 $\pm$ 0.64 & +40.85 \\
8   & 60.55 $\pm$ 1.49 & +40.98 \\
16  & 60.71 $\pm$ 0.83 & +41.14 \\
32  & 59.58 $\pm$ 2.19 & +40.01 \\
64  & 60.79 $\pm$ 2.14 & +41.22 \\
128 & 60.19 $\pm$ 2.01 & +40.62 \\
256 & 59.19 $\pm$ 0.87 & +39.62 \\
\midrule
\multicolumn{3}{l}{\textit{Permutation-invariant baselines (from \cref{tab:routing_ablation})}} \\
GAP   & 19.57 & -- \\
CLS   & 29.01 & -- \\
\bottomrule
\end{tabular}
\end{table}

%% file: appendix/validation-and-reproducibility.tex
\section{Validation and Reproducibility Details}
\label{sec:appendix_validation_details}

\subsection{Eigenvalue Spectrum of the Trained S4 Decoder}
\label{sec:appendix_validation}
\label{sec:app-validation-eigenvalue}

\paragraph{Protocol.} We extract the trained diagonal $\bar{A}$ and compute its 128 complex discrete modes $\{\lambda_i\}$. We report the minimum, maximum, and mean $|\lambda_i|$, the number of slow modes near the unit circle, and the position-wise recurrent coefficient norm $\|\alpha_k\|$ where $\alpha_k = W \cdot \Re(\sum_i C_i\lambda_i^{N-k} \bar{B}_i) + \mathbf{1}_{\{k=N\}}W\mathbf{D}$ for $k\in\{1,\ldots,N\}$ under the final-emitted-state notation. Here $C_i$ is the output coefficient of mode $i$ and $\bar{B}_i$ its discretized input coefficient. We also audit the learned $D$ skip term by evaluating checkpoints with and without the skip path.

\paragraph{Results.} The trained S4 decoders have stable discrete modes. On the five MAE/ImageNet-1K Sinkhorn checkpoints, the dominant mode satisfies $\max_i |\lambda_i| \approx 0.990$ while many modes decay faster (mean $|\lambda_i| \approx 0.306$). This gives a stable LTI kernel combining slow long-memory modes and faster attenuation.

\subsubsection{Post-Hoc Router Interventions}
\label{sec:app-router-posthoc}

\paragraph{Protocol.} We evaluate frozen Sinkhorn checkpoints on 2,048 ImageNet validation images (main seeds 42--46) and 1,024 CUB test images. Sequence interventions include reversal, shuffling, fixed random raw-token order, and ordering by CLS-patch cosine similarity or a random scorer. Cross-seed reuse feeds one scorer's routed sequence to another checkpoint's S4/classifier. Token removal selects the top or bottom 20 patches by Sinkhorn score, CLS cosine, or token norm, alongside random removal.

\begin{table}[ht]
\centering
\caption{Post-hoc router interventions. ImageNet uses five MAE checkpoints on 2,048 images; CUB uses one MAE/CUB checkpoint and two ImageNet-trained routers on 1,024 images. Changes are percentage points; Spearman coefficients and accuracy rows use their stated units.}
\label{tab:posthoc-router-validation}
\begin{tabular}{lcc}
\toprule
Quantity & MAE/ImageNet & MAE/CUB \\
\midrule
Own learned order acc. & $71.18\pm0.36$ & $55.86$ \\
Alternate router order acc. & $71.21\pm0.30$ & $41.21$ \\
Shuffle after routing drop & $-52.14\pm0.31$ & $-43.75$ \\
Random raw-order drop & $-50.46\pm0.41$ & $-50.10$ \\
Drop top-20 Sinkhorn tokens & $-16.34\pm0.75$ & $-41.89$ \\
Drop top-20 CLS-cosine tokens & $-2.02\pm0.14$ & $-8.69$ \\
Drop random 20 tokens & $-0.41\pm0.34$ & $-1.66$ \\
Sinkhorn vs. CLS-cosine Spearman & $-0.04\pm0.01$ & $-0.20$ \\
Sinkhorn vs. token-norm Spearman & $-0.19\pm0.00$ & $+0.01$ \\
\bottomrule
\end{tabular}
\end{table}

\paragraph{Results.} Cross-seed router replacement changes ImageNet accuracy by +.03 points on average, with score-rank Spearman .994 and top-20 overlap 94.5\%. ImageNet scorers paired with the CUB decoder reach 41.21\%, compared with 55.86\% for its own scorer. Shuffling the routed sequence reduces accuracy by 52.14 points on ImageNet and 43.75 on CUB. Removing the top 20 Sinkhorn patches produces larger drops than CLS-cosine or random removal (\Cref{tab:posthoc-router-validation}), connecting scorer selection to the trained readout.

\paragraph{512-image perturbations.} On 512 validation images with the same five ImageNet checkpoints, global rearrangements reduce accuracy by 48--66 points, while adjacent swaps change it by $.16\pm.15$ points (\Cref{tab:global-route-perturbations}). Cross-seed Spearman is .994 with top-20 overlap 94.3\%. Sinkhorn scores have Spearman $-.041$ and overlap 15.5\% with CLS cosine, and $-.200$ and 7.5\% with token norm. \Cref{tab:route-removal-512} compares token removal on this subset.

\begin{table}[ht]
\centering
\caption{Routed-sequence perturbations on 512 ImageNet validation images (five-seed mean $\pm$ std). Changes are paired percentage-point differences from clean accuracy.}
\label{tab:global-route-perturbations}
\begin{tabular}{lcc}
\toprule
Sequence & Accuracy (\%) & Change (pp) \\
\midrule
Learned route & $73.87\pm.70$ & -- \\
Reversed & $7.62\pm.49$ & $-66.25\pm.50$ \\
Half-sequence cyclic shift & $23.28\pm.38$ & $-50.59\pm.65$ \\
Four-block shuffle & $25.70\pm.62$ & $-48.16\pm.26$ \\
Full shuffle & $18.20\pm.65$ & $-55.66\pm.85$ \\
Adjacent swaps & $73.71\pm.62$ & $-.16\pm.15$ \\
\bottomrule
\end{tabular}
\end{table}

\begin{table}[ht]
\centering
\caption{Accuracy decrease after removing 20 patches on the same 512 ImageNet images (five-seed mean $\pm$ std, percentage points).}
\label{tab:route-removal-512}
\begin{tabular}{lrrrr}
\toprule
Selection & Sinkhorn & CLS cosine & Token norm & Random \\
\midrule
Decrease (pp) & $19.9\pm1.2$ & $3.1\pm.5$ & $.0\pm.4$ & $.9\pm.6$ \\
\bottomrule
\end{tabular}
\end{table}

\subsubsection{True-Class Logit under Permutation Controls}
\label{sec:app-validation-bruteforce}

\paragraph{Protocol.} We use the seed-44 MAE/ImageNet checkpoint and the first five validation images, selected before inspecting routes. Soft routing and hard permutations use identical source tokens and sequence lengths. The percentile is the fraction of hard permutations whose true-class logit is at most the soft readout's logit. Analytical S4 logits match direct forwards on 32 random permutations per image and length, with maximum absolute error below $5\times10^{-6}$.

\paragraph{Scale 1: $N=9$.} We subsample source-grid rows and columns 5--7 (zero-based indices 75, 76, 77, 89, 90, 91, 103, 104, 105) and enumerate all $9! = 362{,}880$ permutations. This restricted central-region readout has 4/5 correct predictions and a mean percentile of 69.30\%.

\begin{table}[ht]
\centering
\caption{Exact enumeration of hard permutations on nine central source patches. ``Higher logits'' counts permutations exceeding the soft Sinkhorn true-class logit.}
\label{tab:bruteforce}
\begin{tabular}{ccccc}
\toprule
Image & True Class & Correct? & Higher logits & Percentile \\
\midrule
1 & 91 (lizard) & \checkmark & 67,018 / 362,880 & 81.53\% \\
2 & 171 (greyhound) & \texttimes & 109,104 / 362,880 & 69.93\% \\
3 & 980 (volcano) & \checkmark & 157,922 / 362,880 & 56.48\% \\
4 & 218 (Komondor) & \checkmark & 120,960 / 362,880 & 66.67\% \\
5 & 19 (bulbul) & \checkmark & 102,095 / 362,880 & 71.87\% \\
\midrule
\multicolumn{2}{l}{Mean} & 4/5 & 111,419.8 & \textbf{69.30\%} \\
\bottomrule
\end{tabular}
\end{table}

\paragraph{Scale 2: $N=196$.} We sample $7.2 \times 10^6$ random permutations per image ($36 \times 10^6$ total) from the $196!$ space, using sampling seeds 42--46 in image order. The full-grid soft readout has 5/5 correct predictions and an aggregate percentile of 97.43\%.

\begin{table}[ht]
\centering
\caption{Random permutation sampling on full $N=196$ patches (MAE backbone, ImageNet-1K).}
\label{tab:random196}
\begin{tabular}{ccccc}
\toprule
Image & True Class & Correct? & Higher logits & Percentile \\
\midrule
1 & 91 (lizard) & \checkmark & 136,680 / 7.2M & 98.10\% \\
2 & 171 (greyhound) & \checkmark & 85,185 / 7.2M & 98.82\% \\
3 & 980 (volcano) & \checkmark & 177,012 / 7.2M & 97.54\% \\
4 & 218 (Komondor) & \checkmark & 153,158 / 7.2M & 97.87\% \\
5 & 19 (bulbul) & \checkmark & 373,304 / 7.2M & 94.82\% \\
\midrule
\multicolumn{2}{l}{Aggregate (36M total)} & 5/5 & 925,339 / 36M & \textbf{97.43\%} \\
\bottomrule
\end{tabular}
\end{table}

%% file: appendix/route-controls.tex
\section{Semantic Alignment and Decoder Controls}
\label{sec:route-controls}

\paragraph{Evaluation conventions.}
CUB controls use frozen backbones, the official train/test split, batch size 256, and 100 epochs. Five-seed summaries use seeds 42--46 and sample standard deviations. Settings are matched within each training family. The MAE temperature-control family has baseline $61.64\pm1.11\%$, while the main comparison has $56.67\pm7.54\%$ (\Cref{tab:ablation-fine-grained}); each retains its own seed summary. Cars and ImageNet audits specify their image sets below.

\subsection{Foreground alignment}
\label{sec:foreground-control}
We match all 5,794 CUB test images to the official foreground segmentation masks. Each mask undergoes the spatial preprocessing of its corresponding backbone image processor. A patch is foreground when more than half its transformed pixels belong to the bird mask. We select the 20 source patches with the largest expected routed positions. The uniform random-patch baseline measures foreground coverage on the same grid; the location-bias null retains the route's marginal preference for grid locations while removing image-specific selection. \Cref{tab:cub-foreground} reports the resulting foreground fractions. MAE, BEiT, and supervised ViT use $14\times14$ grids (196 patches); DINOv2 uses a $16\times16$ grid (256 patches). The different processors explain the differing random foreground coverage.

\subsection{Fixed-position decoder and seed variation}
Section~\ref{sec:diagnostic-validation} defines the fixed-position readout and reports aggregate decoder comparisons. The fixed-position MAE decoder reaches $46.49\pm18.17\%$ (per-seed accuracies: 22.28, 59.48, 31.53, 58.96, and 60.18\%).

\begin{table}[ht]
\centering\small
\caption{Per-seed top-20 compactness on the same 256 CUB test images, ordered by seeds 42--46. Aggregate summaries appear in \Cref{tab:decoder-geometry-summary}.}
\label{tab:decoder-geometry}
\begin{tabular}{llc}
\toprule
Backbone & Decoder & Per-seed $R_{20}$ \\
\midrule
MAE & Fixed weights & .700, .933, .744, .533, .950 \\
BEiT & Fixed weights & .687, 1.075, 1.069, .606, .591 \\
DINOv2 & Fixed weights & .910, .916, .646, .800, .898 \\
BEiT & S4 & .681, .631, .733, .598, .550 \\
DINOv2 & S4 & .654, .535, .470, .517, .477 \\
\bottomrule
\end{tabular}
\end{table}

Per-seed coefficient flatness and test accuracy show no consistent separation between compact and noncompact BEiT fixed-weight runs.

\subsection{Agreement with backbone attention}
\label{sec:attention-control}
We compare maps on the same first 256 CUB test images across all four backbones. Attention is averaged over heads. For rollout, each layer contributes $\tfrac12 A_\ell+\tfrac12 I$; the product of these matrices supplies the CLS-to-patch row, renormalized over patches. Last-layer attention uses the head-averaged final CLS-to-patch row. Spearman correlation compares route scores with each attention map; top-20 IoU is intersection size divided by union size.

\Cref{tab:route-attention-agreement} reports all four backbones using their full patch grids.

MAE masking is disabled. Its rollout patch variance has mean $1.094\times10^{-6}$ and minimum $1.806\times10^{-7}$, indicating a relatively flat relevance field. Route and attention agreement is reported alongside this within-image variation.

\subsection{Input perturbations}
\label{sec:input-control}
On the same first 128 CUB test images, the MAE/S4 checkpoint yields $59.375\%$ clean accuracy and $57.8125\%$ with Gaussian noise of standard deviation .005 added after preprocessing. Noise uses a dedicated seed-42 generator, shared between accuracy/geometry and rank-agreement audits. Clean/noisy $R_{20}$ is .6702/.6695, and ERSI is .4895/.4883 bits per patch. Clean-to-noisy route Spearman is $.9953$, top-20 overlap (intersection divided by 20) is $96.9141\%$, and normalized expected-position displacement is .0109. Two independent clean forwards give Spearman 1.000, top-20 overlap 100\%, and displacement below $5\times10^{-7}$.

A center crop retaining 90\% of each spatial dimension (81\% of image area) gives Spearman $.6639$, top-20 overlap $62.6563\%$, and displacement .1546 when compared at relative resized-grid positions. For a source-coordinate comparison, we map the cropped route field to 144 common interior source-grid patch centers using bilinear sampling. Within that shared domain, Spearman is $.7832$, top-20 overlap is $69.2969\%$, and displacement is .1241 on the original 196-position route scale. The crop changes image content and preprocessing; the aligned measurements also include interpolation.

\subsection{Source-grid shape descriptors}
\label{sec:corrected-mae-shapes}
On the first 128 test images of each dataset, we select the 20 largest MAE route scores after restoring patch tokens to the source $14\times14$ grid. \Cref{tab:mae-source-shapes} reports complementary geometric descriptors. Connected components use four-neighbor adjacency; hull area is divided by the squared grid span; MST length is divided by $\sqrt{20}$ times the grid span; entropy uses a $4\times4$ spatial binning. The random column averages 2,048 uniform patch subsets with seed 42. The 128-image compactness summaries and the 512-image main geometry audit use distinct image counts.

\begin{table}[ht]
\centering\small
\caption{MAE top-20 source-grid shapes (128 images per dataset).}
\label{tab:mae-source-shapes}
\begin{tabular}{lrrr}
\toprule
Descriptor & CUB & Cars & Random \\
\midrule
Mean pairwise grid distance & 4.910 & 4.010 & 7.321 \\
$R_{20}$ & .671 & .548 & 1.000 \\
Largest component fraction & .588 & .675 & .133 \\
Number of components & 6.547 & 3.477 & 16.482 \\
Convex hull area fraction & .354 & .176 & .729 \\
Normalized MST length & .492 & .383 & .748 \\
Normalized spatial entropy & .632 & .567 & .847 \\
\bottomrule
\end{tabular}
\end{table}

For the fixed seed-selected CUB test image at index 5238, MAE's top-20 source patches form three components, with 18 patches in the largest component. The top-20 score and absolute-contribution sets overlap by 97.23\% on the 128-image CUB subset, linking the score-based descriptors to the coefficient-based readout audit.

\subsection{MAE/ImageNet source-grid geometry}
\label{sec:imagenet-source-geometry}
We evaluate the five main MAE/ImageNet Sinkhorn checkpoints (seeds 42--46) on the same first 512 validation images. Patch tokens are restored with $\texttt{ids\_restore}$ before assigning source-grid coordinates. A seed-42 split assigns 256 routes to calibration and 256 to held-out evaluation; each checkpoint's spatial scale is selected on calibration routes. The random compactness reference averages 2,048 uniform top-20 subsets. Frozen features use bfloat16 and readouts use float32.

\begin{table}[ht]
\centering
\caption{MAE/ImageNet source-grid geometry on 512 validation images. Error bars are sample standard deviations across the five checkpoints.}
\label{tab:imagenet-source-geometry}
\begin{tabular}{lccc}
\toprule
Seed & ERSI (bits/patch) & $\sigma$ & $R_{20}$ \\
\midrule
42 & .3088 & 2 & .5806 \\
43 & .3196 & 2 & .5791 \\
44 & .3073 & 2 & .5834 \\
45 & .3043 & 2 & .5828 \\
46 & .3137 & 2 & .5784 \\
\midrule
Mean $\pm$ std & $.3107\pm.0060$ & 2 & $.5809\pm.0022$ \\
\bottomrule
\end{tabular}
\end{table}

A separate shape audit on the first 128 validation images gives a largest-component fraction of $.571\pm.005$, $5.917\pm.087$ components, hull-area fraction $.241\pm.002$, normalized MST length $.427\pm.001$, and spatial entropy $.630\pm.002$ across seeds. Their random references are .133, 16.482, .729, .748, and .847, respectively. The 128-image compactness ratio is $.575\pm.003$.

\subsection{Source-ordered scan interventions and computation}
\label{sec:corrected-mae-heads}
We evaluate seed-42 final CUB and Cars checkpoints on the first 256 test images. Each MAE multi-direction head averages four S4 final outputs, with each path traversing all 196 source patches. Reversal and shuffling change the input sequence before applying each head's specified traversal. Dropping removes the 20 patches with largest cosine similarity to the frozen CLS feature and filters each original path by the retained source indices. Random-dynamic evaluation uses seed 42 for each condition. These subset accuracies describe within-checkpoint interventions; the five-seed, full-test accuracies appear in \Cref{tab:ablation-fine-grained}.

\begin{table}[ht]
\centering\small
\caption{Source-grid MAE scan interventions. Each cell gives CUB / Cars accuracy (\%).}
\label{tab:mae-source-interventions}
\begin{tabular}{lrrrr}
\toprule
Readout & Clean & Reverse & Shuffle & Drop top-20 \\
\midrule
Raster & 14.45 / 13.67 & 5.86 / 3.91 & 5.47 / 3.12 & 14.45 / 14.84 \\
VMamba-style & 12.89 / 18.36 & 12.89 / 18.36 & 13.28 / 11.72 & 12.50 / 17.97 \\
Diagonal & 19.92 / 23.05 & 19.92 / 23.05 & 14.45 / 16.41 & 19.53 / 19.92 \\
Snake & 18.75 / 17.97 & 11.72 / 16.02 & 11.72 / 8.20 & 17.58 / 17.19 \\
Random-fixed & 11.72 / 19.53 & 8.98 / 8.98 & 7.81 / 12.89 & 10.16 / 16.41 \\
Random-dynamic & 32.81 / 42.97 & 35.55 / 45.31 & 33.98 / 45.31 & 29.69 / 44.14 \\
Bi-GRU (Raster) & 28.12 / 31.25 & 14.06 / 10.94 & 22.66 / 15.62 & 25.00 / 26.95 \\
\bottomrule
\end{tabular}
\end{table}

The VMamba-style and diagonal path sets are closed under sequence reversal, so averaging their four outputs preserves the clean prediction under reversal. Random-dynamic interventions include fresh random traversals and consequently vary with that sampling.

\begin{table}[ht]
\centering\small
\caption{MAE/CUB scan head budgets and CPU latency. Batch size 1, float32, four CPU threads, Intel Xeon Gold 6530, PyTorch 2.5.1; median of 30 forwards after five warmups. Latencies include traversal and head computation, with the frozen backbone excluded.}
\label{tab:mae-source-head-cost}
\begin{tabular}{lrrr}
\toprule
Readout & Parameters & Patch visits & Median (ms) \\
\midrule
Raster & 351,689 & 196 & 9.44 \\
VMamba-style & 351,689 & 784 & 35.57 \\
Diagonal & 351,689 & 784 & 34.81 \\
Snake & 351,689 & 784 & 34.49 \\
Random-fixed & 351,689 & 196 & 9.50 \\
Random-dynamic & 351,689 & 196 & 7.32 \\
Bi-GRU (Raster) & 741,064 & 196 & 10.22 \\
\bottomrule
\end{tabular}
\end{table}

Four-direction heads share their parameters across paths and process four times the single-scan patch visits. CUB heads have 351,689 parameters and Cars heads 348,613. MAE/BEiT/ViT make 784 patch visits per image; DINOv2 makes 1,024. \Cref{tab:probe-complexity-full} reports forward-operation estimates.

\begin{table}[ht]
\centering\small
\caption{Full-grid scan CPU latency (ms), shown as CUB / Cars. Batch 1, float32, four CPU threads, Intel Xeon Gold 6530, PyTorch 2.5.1. Each value is the median of 30 forwards after five warmups, with seed-42 randomized interleaving across the 18 heads. Frozen backbone computation is excluded.}
\label{tab:fullscan-head-cost}
\begin{tabular}{lrrr}
\toprule
Backbone & VMamba-style & Diagonal & Snake \\
\midrule
BEiT & 39.66 / 31.18 & 39.44 / 30.92 & 39.32 / 31.17 \\
ViT & 50.69 / 50.85 & 49.87 / 49.18 & 50.50 / 46.39 \\
DINOv2 & 46.98 / 49.80 & 48.30 / 53.05 & 46.84 / 48.78 \\
\bottomrule
\end{tabular}
\end{table}

\begin{table}[ht]
\centering\small
\caption{Complete-path scan interventions on the first 256 test images per dataset. Each cell gives CUB / Cars accuracy (\%). Seed-42 final checkpoints use the full source grid. Dropping removes the 20 source patches with largest CLS cosine similarity from each traversal, preserving the order of the remaining patches.}
\label{tab:fullscan-interventions}
\begin{tabular}{lrrrr}
\toprule
Readout & Clean & Reverse & Shuffle & Drop top-20 \\
\midrule
BEiT / VMamba & 82.42 / 65.62 & 82.42 / 65.62 & 87.89 / 63.67 & 81.25 / 66.02 \\
BEiT / Diagonal & 82.03 / 68.36 & 82.03 / 68.36 & 87.11 / 66.41 & 79.69 / 67.19 \\
BEiT / Snake & 82.03 / 66.41 & 83.20 / 65.23 & 87.50 / 66.80 & 80.08 / 66.80 \\
ViT / VMamba & 80.86 / 55.08 & 80.86 / 55.08 & 81.25 / 52.73 & 80.47 / 56.64 \\
ViT / Diagonal & 82.42 / 56.25 & 82.42 / 56.25 & 82.81 / 56.64 & 80.47 / 57.03 \\
ViT / Snake & 81.64 / 57.81 & 82.03 / 54.30 & 84.38 / 51.95 & 80.86 / 58.20 \\
DINOv2 / VMamba & 63.67 / 43.75 & 63.67 / 43.75 & 56.25 / 32.42 & 62.89 / 42.58 \\
DINOv2 / Diagonal & 59.77 / 46.88 & 59.77 / 46.88 & 54.69 / 28.52 & 58.20 / 46.09 \\
DINOv2 / Snake & 63.67 / 48.44 & 57.81 / 32.42 & 43.75 / 19.92 & 61.72 / 49.22 \\
\bottomrule
\end{tabular}
\end{table}

VMamba-style and diagonal path sets preserve their accuracy under sequence reversal. Shuffling changes the source-grid adjacency encountered by the fixed traversals; token removal changes both the available evidence and the effective sequence length. These subset measurements characterize each checkpoint separately from the full-test five-seed summaries.

\subsection{Routing hyperparameters and matched readouts}
\begin{table}[ht]
\centering
\caption{CUB accuracy controls (five-seed mean $\pm$ sample std). Temperature and iteration settings are matched within each backbone's sweep. Scorer weight decay uses a separate matched pair.}
\label{tab:routing-controls}
\begin{tabular}{llc}
\toprule
Backbone & Configuration & Accuracy (\%) \\
\midrule
DINOv2 & $\tau=.05$, 20 iterations & $80.64\pm1.79$ \\
DINOv2 & $\tau=.10$, 5 iterations & $81.22\pm3.22$ \\
DINOv2 & $\tau=.10$, 20 iterations & $81.19\pm3.67$ \\
DINOv2 & $\tau=.20$, 20 iterations & $82.72\pm1.07$ \\
MAE & $\tau=.10$, 20 iterations & $61.64\pm1.11$ \\
MAE & $\tau=.20$, 20 iterations & $59.76\pm.78$ \\
\midrule
DINOv2 & Scorer weight decay 0 & $81.86\pm1.23$ \\
DINOv2 & Scorer weight decay $10^{-3}$ & $81.40\pm3.01$ \\
\bottomrule
\end{tabular}
\end{table}
These sweeps report accuracy sensitivity; geometry sensitivity to routing hyperparameters is listed in Section~\ref{sec:limitations}.

\Cref{tab:matched-dinov2-readouts} reports the matched DINOv2/CUB readout family under a common training budget.

\begin{table}[ht]
\centering
\caption{Matched DINOv2/CUB readouts from the independent control family (five-seed mean $\pm$ sample std). Complete four-path scan results appear in \Cref{tab:ablation-fine-grained}.}
\label{tab:matched-dinov2-readouts}
\begin{tabular}{lc}
\toprule
Readout & Accuracy (\%) \\
\midrule
CLS & $89.45\pm.21$ \\
GAP & $68.53\pm.24$ \\
Sinkhorn/S4 & $81.17\pm2.03$ \\
Larger Transformer & $87.15\pm.17$ \\
DeepSets & $76.93\pm.17$ \\
Sinkhorn/1D-convolution & $71.01\pm.64$ \\
\midrule
Raster & $35.36\pm.31$ \\
Random-dynamic & $69.04\pm.09$ \\
Random-fixed & $38.40\pm7.49$ \\
Bi-GRU & $59.64\pm1.57$ \\
\bottomrule
\end{tabular}
\end{table}

%% file: checklist.tex
\section*{NeurIPS Paper Checklist}

\begin{enumerate}

\item {\bf Claims}
    \item[] Question: Do the main claims made in the abstract and introduction accurately reflect the paper's contributions and scope?
    \item[] Answer: \answerYes{}
    \item[] Justification: The abstract and introduction state that SSMProbe is an inspectable Sinkhorn+S4 diagnostic for evidence routing, and that route geometry differs across frozen ViT pre-training objectives. These claims are supported by the ImageNet-1K MAE results in \Cref{tab:mae-main}, the cross-backbone fine-grained results in \Cref{tab:ablation-fine-grained}, and the evidence-route geometry summary in \Cref{tab:diagnostic-summary}.

\item {\bf Limitations}
    \item[] Question: Does the paper discuss the limitations of the work performed by the authors?
    \item[] Answer: \answerYes{}
    \item[] Justification: \Cref{sec:limitations} discusses task and decoder dependence, evaluation scope, and input sensitivity.

\item {\bf Theory assumptions and proofs}
    \item[] Question: For each theoretical result, does the paper provide the full set of assumptions and a complete (and correct) proof?
    \item[] Answer: \answerYes{}
    \item[] Justification: The appendix provides formal derivations of the LTI dynamics, position-dependent modulation mechanism, and the information scheduling interpretation of the learned Sinkhorn permutation. Assumptions (discrete LTI, causal scanning, permutation sensitivity) are explicitly stated.

\item {\bf Experimental result reproducibility}
    \item[] Question: Does the paper fully disclose all the information needed to reproduce the main experimental results of the paper to the extent that it affects the main claims and/or conclusions of the paper (regardless of whether the code and data are provided or not)?
    \item[] Answer: \answerYes{}
    \item[] Justification: \Cref{sec:experimental-setup} specifies the backbones, datasets, probe families, optimization (AdamW, lr=$10^{-3}$, cosine schedule, batch size 256), and epoch counts. Appendix~\ref{sec:appendix_implementation} provides S4 and Sinkhorn implementation details.

\item {\bf Open access to data and code}
    \item[] Question: Does the paper provide open access to the data and code, with sufficient instructions to faithfully reproduce the main experimental results, as described in supplemental material?
    \item[] Answer: \answerNo{}
    \item[] Justification: Code and instructions for reproducing the main experimental results will be released publicly after the camera-ready submission. ImageNet-1K is a standard benchmark with access subject to its terms. Experiment configurations are detailed in the Experimental Setup section.

\item {\bf Experimental setting/details}
    \item[] Question: Does the paper specify all the training and test details (e.g., data splits, hyperparameters, how they were chosen, type of optimizer) necessary to understand the results?
    \item[] Answer: \answerYes{}
    \item[] Justification: \Cref{sec:experimental-setup} provides hyperparameter specifications including backbone, dataset, probe head families, optimizer type (AdamW), learning rate ($10^{-3}$), schedule (cosine), batch size (256), and number of epochs (5 for ImageNet-1K, 100 for CUB/Cars).

\item {\bf Experiment statistical significance}
    \item[] Question: Does the paper report error bars suitably and correctly defined or other appropriate information about the statistical significance of the experiments?
    \item[] Answer: \answerYes{}
    \item[] Justification: Multi-seed evaluations use the fixed seed set \{42, 43, 44, 45, 46\}. Summaries report mean and standard deviation with the number of valid runs and numerical failures stated explicitly. The DINOv2/CUB full-grid VMamba-style scan at seed 44 develops NaN loss during epoch 96 of 100. This run is retained as a numerical failure without replacement or protocol changes; its final NaN-output accuracy is excluded from valid-run accuracy summaries. Random-fixed permutations also exhibit substantial seed variation.

\item {\bf Experiments compute resources}
    \item[] Question: For each experiment, does the paper provide sufficient information on the computer resources (type of compute workers, memory, time of execution) needed to reproduce the experiments?
    \item[] Answer: \answerYes{}
    \item[] Justification: Experiments use a single H100 80GB GPU. Seven-head MAE training shares the frozen backbone forward and takes 19.8--20.2 minutes per CUB seed, 36.5--39.2 minutes per Cars seed (100 epochs), and 2.49--3.26 hours per ImageNet seed (5 epochs). The BEiT/ViT/DINOv2 complete-path controls run as three concurrent jobs, with three heads per job: CUB takes 18.8--24.6 minutes per seed and Cars 37.4--45.9 minutes. Timings include final test evaluation. Combined GPU memory during the concurrent jobs was observed at approximately 32 GiB; this is a spot measurement. Frozen backbones use bfloat16 and heads use float32. Appendix~\ref{sec:appendix_complexity} reports probe parameters and forward-operation estimates.

\item {\bf Code of ethics}
    \item[] Question: Does the research conducted in the paper conform, in every respect, with the NeurIPS Code of Ethics?
    \item[] Answer: \answerYes{}
    \item[] Justification: This work presents fundamental research on representation probing without direct societal applications. All experiments use established benchmarks and publicly available model checkpoints.

\item {\bf Broader impacts}
    \item[] Question: Does the paper discuss both potential positive societal impacts and negative societal impacts of the work performed?
    \item[] Answer: \answerNA{}
    \item[] Justification: This is foundational representation learning research focused on probing methodology. It does not propose new applications with direct societal impact.

\item {\bf Safeguards}
    \item[] Question: Does the paper describe safeguards that have been put in place for responsible release of data or models that have a high risk for misuse?
    \item[] Answer: \answerNA{}
    \item[] Justification: This work uses standard pretrained models and benchmark datasets without release risks. No high-risk dual-use models are involved.

\item {\bf Licenses for existing assets}
    \item[] Question: Are the creators or original owners of assets (e.g., code, data, models), used in the paper, properly credited and are the license and terms of use explicitly mentioned and properly respected?
    \item[] Answer: \answerYes{}
    \item[] Justification: ViT-MAE-base is credited to He et al. (2022) under Apache 2.0. ImageNet-1K is cited as the standard benchmark. S4 implementation is credited to Gu et al. (2021). DINOv2 is credited to Oquab et al. (2023/2024). CUB-200-2011 (Caltech-UCSD Birds-200-2011) is cited as a standard fine-grained visual classification benchmark. Stanford Cars is cited as a standard fine-grained vehicle classification benchmark.

\end{enumerate}